\documentclass[runningheads]{llncs}

\usepackage{eccv}

\usepackage{eccvabbrv}

\usepackage{graphicx}
\usepackage[width=122mm,left=12mm,paperwidth=146mm,height=193mm,top=12mm,paperheight=217mm]{geometry}
\usepackage{booktabs}

\usepackage[accsupp]{axessibility}  

\usepackage[pagebackref,breaklinks,colorlinks,citecolor=eccvblue]{hyperref}

\usepackage{orcidlink}

\usepackage{paralist}
\usepackage{pifont}
\usepackage{tabu}
\usepackage{diagbox}
\usepackage{multirow} 

\usepackage{colortbl}
\definecolor{gray}{rgb}{0.3,0.3,0.3}
\definecolor{blue}{rgb}{0,0.5,1}
\newcommand{\green}[1]{\textcolor[RGB]{96,177,87}{#1}}
\newcommand{\gbf}[1]{\green{#1}}

\newcommand{\cmark}{\ding{51}}%
\newcommand{\xmark}{\ding{55}}%
\usepackage{wrapfig}

\makeatletter
\def\thanks#1{\protected@xdef\@thanks{\@thanks
        \protect\footnotetext{#1}}}
\makeatother

\begin{document}

\title{Open Panoramic Segmentation} 


\author{Junwei Zheng\inst{1}\orcidlink{0009-0005-4390-3044} \and
Ruiping Liu\inst{1}\orcidlink{0000-0001-5245-2277} \and
Yufan Chen\inst{1}\orcidlink{0009-0008-3670-4567}, \\ Kunyu Peng\inst{1}\orcidlink{0000-0002-5419-9292} \and Chengzhi Wu\inst{1}\orcidlink{0000-0003-2186-3748} \and Kailun Yang\inst{2}\orcidlink{0000-0002-1090-667X}, \\ Jiaming Zhang\inst{1,\dag}\orcidlink{0000-0003-3471-328X} \and Rainer Stiefelhagen\inst{1}\orcidlink{0000-0001-8046-4945}}\thanks{$^{\dag}$~Correspondence: jiaming.zhang@kit.edu}

\authorrunning{J.~Zheng et al.}

\institute{Karlsruhe Institute of Technology \and Hunan University}

\maketitle

\begin{abstract}
Panoramic images, capturing a 360{\textdegree} field of view (FoV), encompass omnidirectional spatial information crucial for scene understanding. However, it is not only costly to obtain training-sufficient dense-annotated panoramas but also application-restricted when training models in a close-vocabulary setting. To tackle this problem, in this work, we define a new task termed \textbf{Open Panoramic Segmentation (OPS)}, where models are trained with FoV-restricted pinhole images in the source domain in an open-vocabulary setting while evaluated with FoV-open panoramic images in the target domain, enabling the zero-shot open panoramic semantic segmentation ability of models. Moreover, we propose a model named OOOPS with a Deformable Adapter Network (DAN), which significantly improves zero-shot panoramic semantic segmentation performance. To further enhance the distortion-aware modeling ability from the pinhole source domain, we propose a novel data augmentation method called Random Equirectangular Projection (RERP) which is specifically designed to address object deformations in advance. Surpassing other state-of-the-art open-vocabulary semantic segmentation approaches, a remarkable performance boost on three panoramic datasets, WildPASS, Stanford2D3D, and Matterport3D, proves the effectiveness of our proposed OOOPS model with RERP on the OPS task, especially $\textbf{{+}2.2\%}$ on outdoor WildPASS and $\textbf{{+}2.4\%}$ mIoU on indoor Stanford2D3D. The source code is publicly available at \href{https://junweizheng93.github.io/publications/OPS/OPS.html}{OPS}.
\end{abstract}

\section{Introduction}
\label{sec:intro}
\begin{figure}[t!]
    \centering
    \begin{minipage}{0.5\textwidth}
        \centering
        \begin{subfigure}[t]{0.98\columnwidth}
            \centering
            \includegraphics[width=1.0\linewidth,height=1.9cm]{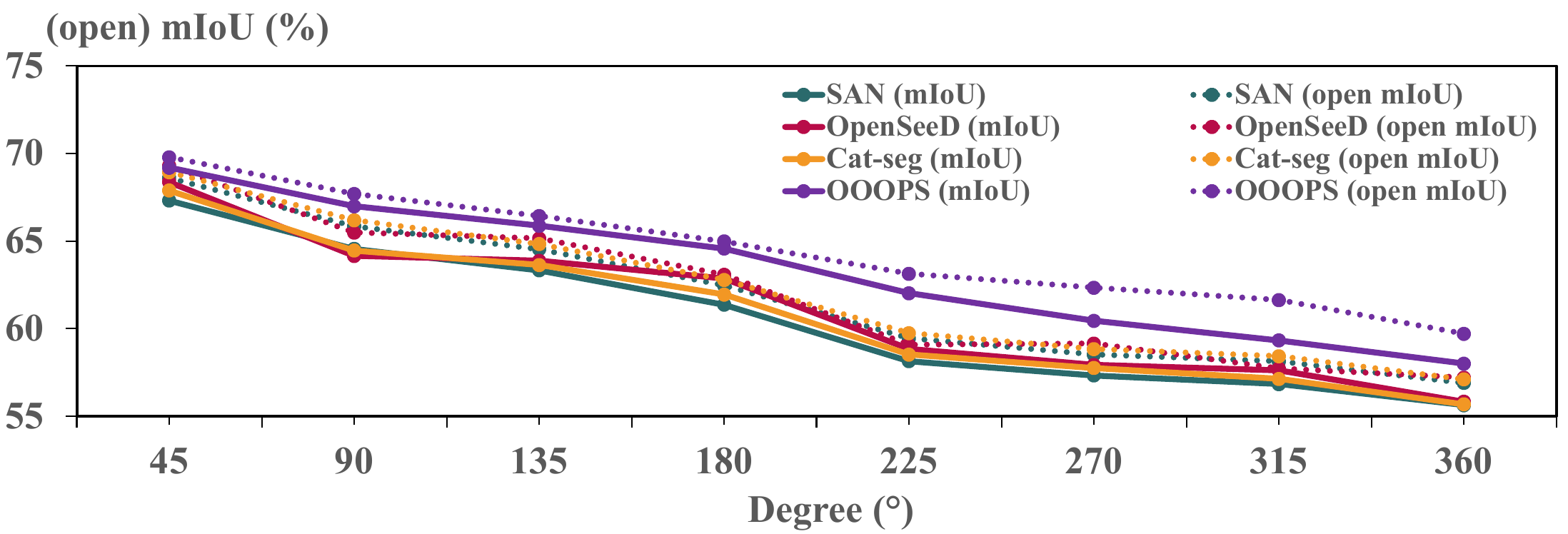}
            \caption{The challenge of wider FoV. Performance (mIoU / open mIoU) drops when FoV increases.
            }
            \label{fig1-a:performance_drop}
        \end{subfigure}
        \begin{subfigure}[t]{0.98\columnwidth}
            \centering
            \includegraphics[width=1.0\linewidth,height=2.1cm
            ]{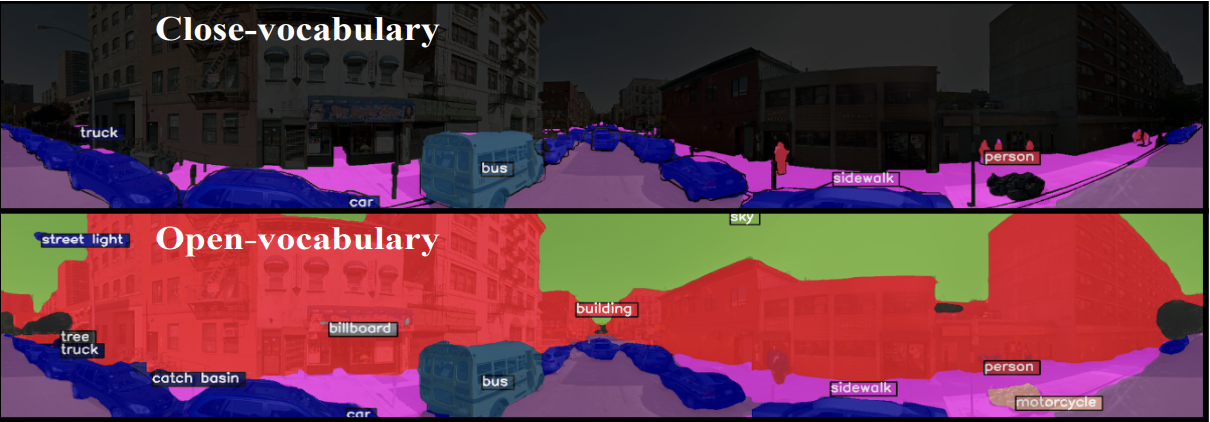}
            \caption{Close-vocabulary \textit{vs.} Open-vocabulary segmentation on panoramas. While limited categories are recognized in a close-vocabulary setting, all objects in 360{\textdegree} can be covered in an open-vocabulary setting. 
            }
            \label{fig1-b:close_vs_open_pano_seg}
        \end{subfigure}
    \end{minipage}%
    \begin{minipage}{0.5\textwidth}
        \centering
        \begin{subfigure}[t]{0.98\columnwidth}
            \includegraphics[width=1.0\linewidth,height=5.3cm
            ]{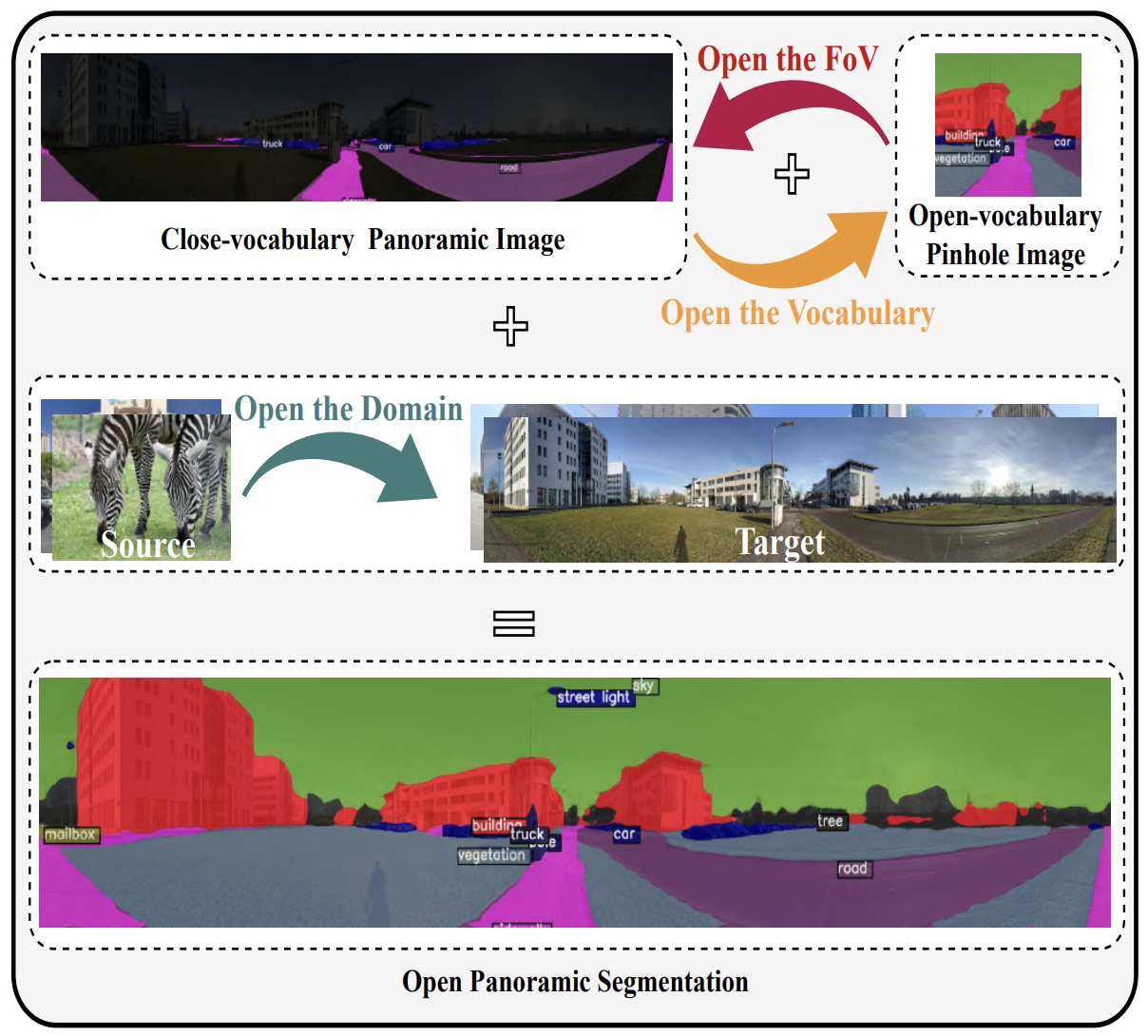}
            \caption{The paradigm of \textbf{Open Panoramic Segmentation (OPS)} task.}
            \label{fig1-c:ops_task}
        \end{subfigure}
    \end{minipage}
    \caption{(a) The challenge of existing state-of-the-art segmentation models. 
    (b) The limitation of categories in traditional close-vocabulary panoramic segmentation tasks.
    (c) Our newly defined \textbf{Open Panoramic Segmentation (OPS)} task aims at tackling the above challenges. OPS consists of three important elements: \textbf{\textcolor[RGB] {184,13,72}{Open the FoV}}  targeted at the challenge of 360\textdegree~FoV, \textbf{\textcolor[RGB]{242,151,36}{Open the Vocabulary}} targeted at the drawback of close-vocabulary panoramic segmentation and \textbf{\textcolor[RGB]{60,124,124}{Open the Domain}} targeted at the challenge of scarcity of panoramic labels.}
    \label{fig1:challenge_and_ops_task}
\end{figure}
Panoramic imaging systems~\cite{jiang2022annular,jiang2023minimalist} have advanced significantly in recent years, which has fostered a wide variety of panoramic vision applications~\cite{ai2022deep_omnidirectional,gao2022review}. 
Due to the comprehensive 360{\textdegree} Field of View (FoV), omnidirectional panoramas provide more informative visual cues when perceiving surroundings in a broad spectrum of scene understanding tasks~\cite{ling2023panoswin,teng2023360bev,shi2023panoflow,athwale2023darswin,yu2023panelnet,shen2022panoformer}, enabling a more complete and immersive capture of environmental data, which is crucial for in-depth scene understanding.
This wide perspective surpasses the limited scope of pinhole images, significantly enhancing the capability of computer vision systems to perceive and interpret surroundings in a variety of applications.
While the benefits of utilizing panoramic images in computer vision applications are apparent compared with pinhole images, it is imperative to give increasing consideration to some noteworthy challenges as follows: 
(1) \emph{The challenge of wider FoV}. For example, Fig.~\ref{fig1-a:performance_drop} illustrates the degraded performance of state-of-the-art open-vocabulary semantic segmentation approaches~\cite{xu2023san, cho2023cat_seg, zhang2023openseed} with an increasing FoV, \ie, from pinhole images to 360\textdegree~panoramic images. 
Over $12\%$ mIoU performance degradation can be observed, indicating the challenges introduced by the large divergence of semantic and structural information across narrow-wide imagery.
(2) \emph{The limitation of categories.} The traditional close-vocabulary segmentation task paradigm~\cite{yang2020omnisupervised, zhang2022trans4pass} only provides a limited amount of labeled categories, which cannot handle the inestimable amount of categories in real-world applications.
Fig.~\ref{fig1-b:close_vs_open_pano_seg} illustrates the difference between close- and open-vocabulary segmentation.
Compared to the close-vocabulary setting (first row) the open-vocabulary setting (second row) is not limited by the number of categories of the datasets.
Only four predefined categories (highlighted in different colors) are recognized in the close-vocabulary setting in contrast to the open one, where every pixel of the panorama has its own semantic meaning even though the categories are not annotated in the dataset.\\
\indent To further release the great potential of panoramic images three critical problems are eager to be solved: \textit{\textbf{\textcolor[RGB]{184,13,72}{How}} to obtain a holistic perception via a single image? 
\textbf{\textcolor[RGB]{242,151,36}{How}} to break the barrier of limited recognizable categories in existing panoramic datasets so that downstream vision applications can benefit from unrestricted informative visual hints? \textbf{\textcolor[RGB]{60,124,124}{How}} to deal with the scarcity of panoramic labels?
}
Based on the aforementioned three questions, we propose a new task named \textbf{Open Panoramic Segmentation (OPS)} that comprehensively addresses these challenges, aiming to better leverage the advantages brought by panoramic images.
The new task paradigm is shown in Fig.~\ref{fig1-c:ops_task}.
The open panoramic segmentation task consists of three important elements considering the three problems.
\textbf{The concept of ``Open'' is three-fold}: 
omnidirectional panoramic images (\textbf{\textcolor[RGB]{184,13,72}{Open the FoV}}), an unrestricted range of recognizable classes (\textbf{\textcolor[RGB]{242,151,36}{Open the Vocabulary}}), and additionally, models are trained using pinhole images in the source domain while evaluated with panoramic images in the target domain (\textbf{\textcolor[RGB]{60,124,124}{Open the Domain}}).
Given that densely annotated pinhole segmentation labels incur lower costs compared to panoramic ones~\cite{yang2020omnisupervised}, it is cost-efficient to open different domains.
Note that OPS is different from Domain Adaptation (DA).
In the context of training in DA, the utilization of data encompasses both the source and target domains while in OPS, exclusive reliance is placed on data originating solely from the source domain during the whole training process.

Apart from the new task, we propose a new model named OOOPS to address the aforementioned three openness-related challenges in the OPS task.
It consists of a frozen CLIP model~\cite{radford2021clip} and a key component named \emph{Deformable Adapter Network (DAN)}, which serves two critical purposes: (1) efficiently adapting the frozen CLIP model to the panoramic segmentation task and (2) addressing object deformation and image distortion inherent in panoramas.
More specifically, a novel Deformable Adapter Operator (DAO), the key component of DAN, is designed to tackle panoramic distortion.
To advance the modeling capabilities for distortion awareness in the pinhole source domain, we further introduce \emph{Random Equirectangular Projection (RERP)} explicitly crafted to tackle object deformations and image distortions.
A pinhole image is divided into four image patches that are randomly shuffled.
Equirectangular Projection, one of the most common methods for mapping a globe into a panoramic plane, is then introduced in the shuffled image.
Our OOOPS model with RERP outperforms other state-of-the-art open-vocabulary segmentation methods with $\textbf{{+}2.2\%}$, $\textbf{{+}2.4\%}$, and $\textbf{{+}0.6\%}$ mIoU on WildPASS~\cite{yang2021wildpass}, Stanford2D3D~\cite{armeni1702s2d3d}, and Matterport3D~\cite{Matterport3D}, respectively.

To summarize, we present the following contributions:
\begin{compactitem}
\item We introduce a new task termed open panoramic segmentation, \textbf{OPS} for short, including \textbf{Open FoV}, \textbf{Open Vocabulary} and \textbf{Open Domain}.
Models are trained with FoV-restricted pinhole images in the source domain in an open-vocabulary setting while evaluated with FoV-open panoramic images in the target domain.
\item We put forward a model called OOOPS with the aim to address three openness-related challenges at once. A \textbf{Deformable Adapter Network (DAN)} is proposed to
transfer the zero-shot learning ability of a frozen CLIP model from the pinhole domain to different panoramic domains. 
\item A novel data augmentation strategy named \textbf{Random Equirectangular Projection (RERP)}, which is specifically designed for the proposed OPS task, further boosts the accuracy of the OOOPS model, achieving state-of-the-art performance in the open panoramic segmentation task. 
\item For benchmarking OPS, we conduct comprehensive evaluations on both indoor and outdoor datasets (WildPASS, Stanford2D3D and Matterport3D), involving more than 10 close- and open-vocabulary segmentation models.   
\end{compactitem}

\section{Related Work}
\label{sec:related_work}
\subsection{Open-vocabulary Semantic Segmentation}
Motivated by the success of vision-language pre-training models, Open-Vocabulary Semantic Segmentation (OVSS) has been recently explored to achieve open-set pixel-wise understanding~\cite{li2022language_driven,xu2022simple_baseline,ghiasi2022scaling, Ding2022OpenVocabularyUI, Takmaz2023OpenMask3DO3, Xu2023MasQCLIPFO, Wang2023HierarchicalOU, Mukhoti2022OpenVS, Li2023OpenvocabularyOS, Ma2022OpenvocabularySS, Guo2023MVPSEGMP, Ma2023OpenVocabularySS, Wysoczanska2023CLIPDIYCD, Li2023TagCLIPID, Chen2023ExploringOS, Dao2023ClassEL, Wang2023DiffusionMI, Li2023OpenvocabularyOS, Wei2023OVPARTSTO,xie2023sed,jiayun2023plug,liu2024multigrained}.
Most works leverage the open-vocabulary capabilities of CLIP~\cite{radford2021clip} to reason about relations between dense pixel semantics and arbitrary class labels~\cite{cho2023cat_seg, Li2023TagCLIPID, Wysoczanska2023CLIPDIYCD, Luo2022SegCLIPPA, Xu2023MasQCLIPFO, Ding2022OpenVocabularyUI}.
In~\cite{zhou2023lmseg,yin2022devil_labels_sentences}, multiple datasets are unified by transferring heterogeneous labels into text embeddings for training to achieve OVSS.
In~\cite{liang2022mask_adapted_clip,xu2023learning_natural}, image-caption datasets are employed for learning alignment between visual regions and text entities. 
To reduce the computational costs of using a frozen pre-trained CLIP model for identifying novel classes, latest studies~\cite{xu2023side_adapter_network,han2023global_knowlegde_calibration} have also proposed efficient architectures for fast OVSS.
SAN~\cite{xu2023side_adapter_network,xu2023san} has designed an asymmetric input resolution strategy to decouple mask proposal prediction and attention bias prediction applied to CLIP for mask class recognition. 
Furthermore, open-vocabulary panoptic segmentation has been tackled in~\cite{Ding2022OpenVocabularyUI,xu2023text_to_image_diffusion,chen2023embedding_modulation,dong2023vloss,zhang2023openseed,qin2023freeseg,Xu2023MasQCLIPFO, xu2023text_to_image_diffusion, Li2023OpenvocabularyOS} to render universal image understanding. 
In this work, we focus on the open-vocabulary zero-shot learning on \textbf{panoramas} instead of pinhole images discussed in the aforementioned works.

\subsection{Panoramic Semantic Segmentation}

Panoramic semantic segmentation enables more informative decision-making and interaction within complex and dynamic environments, advancing the capabilities of intelligent systems for navigation and perception~\cite{yang2021context,jaus2021panoramic_towards,jaus2023panoramic_insights, Li2023SGAT4PASSSG, Orhan2021SemanticSO, BerenguelBaeta2022FreDSNetJM, Jiang2018DFNetSS,fu2023panopticnerf, mei2022waymo}.
PASS~\cite{yang2020pass,yang2019can} achieves panoramic segmentation on annular images, whereas 360BEV~\cite{teng2023360bev} tackles panoramic semantic mapping.
ECANets~\cite{yang2021wildpass} address omnidirectional semantic segmentation with multi-source omni-supervised learning to enhance generalization. 
Zhang~\etal~\cite{zhang2021transfer_beyond} propose to use attention-augmented designs for panoramic representation learning. 
Guttikonda~\etal~\cite{guttikonda2023single_spherical} propose a single-frame panoramic semantic segmentation approach using multi-modal spherical images.
Hu~\etal~\cite{Hu2022DistortionCM} propose a distortion convolution module for panoramic semantic segmentation.
An important cluster of works revisits panoramic semantic segmentation from an unsupervised domain adaptation perspective~\cite{zhang2021transfer_beyond,jang2022dada, zheng2023look_neighbor, ma2021densepass,zheng2023both_style_distortion, kim2022pasts,zhang2022trans4pass}, by adapting from label-rich pinhole images to the panoramic images. 
Unlike these works, we propose a method to tackle \textbf{open-vocabulary} panoramic semantic segmentation for unconstrained surrounding understanding.

\subsection{Deformable Neural Networks}

In the realm of deformable neural networks, notable advancements have been made through various iterations. The pioneering work~\cite{dai2017DCNv1} introduces the concept of deformable convolution, laying the foundation for subsequent developments.
Building upon this foundation, DCNv2~\cite{zhu2019DCNv2} further refines deformable convolution by introducing a modulation mechanism that expands the scope of deformation modeling.
The evolution of deformable neural networks continues with the introduction of DCNv3~\cite{wang2023DCNv3}, which combines the group convolution with DCNv2 for the sake of imitating multi-head mechanism of Transformer~\cite{vaswani2017attention}.
DCNv4~\cite{xiong2024DCNv4} aims to speedup DCNv3, achieving a more efficient model.
Additionally, Trans4PASS~\cite{zhang2022trans4pass,trans4passplus} proposes Deformable Patch Embedding (DPE) and Deformable MLP (DMLP) to handle the distortion that occurs in panoramas.
Incorporating the principles of deformable convolution into vision transformers~\cite{dosovitskiy2020image}, 
Xia~\etal propose DeformableViT~\cite{xia2022DeformableViT,xia2023dat++}, extending the applicability of deformable neural networks to the domain of transformer architectures.
After rethinking a series of DCN designs, we propose \textbf{DAO} to improve the model's ability to handle image distortion in the OPS task.

\section{Methodology}
\label{sec:methodology}
\subsection{Open Panoramic Segmentation}
\label{subsec:ops}
Open Panoramic Segmentation (OPS) task is proposed to deal with three challenging problems: \ding{182} the narrow FoV, \ding{183} the restricted range of classes, and \ding{184} the scarcity of panoramic labels.
OPS gives three answers to the aforementioned issues: \ding{182} Open the FoV, \ding{183} Open the Vocabulary, and \ding{184} Open the Domain.
The OPS task paradigm is shown in Fig.~\ref{fig1-c:ops_task}.
Models are trained in the narrow-FoV pinhole source domain in an open-vocabulary setting while evaluated in the wide-FoV panoramic target domain.

\begin{figure*}[h]
    \centering
    \includegraphics[width=0.99\linewidth,height=6cm,trim=2 2 2 2,clip]{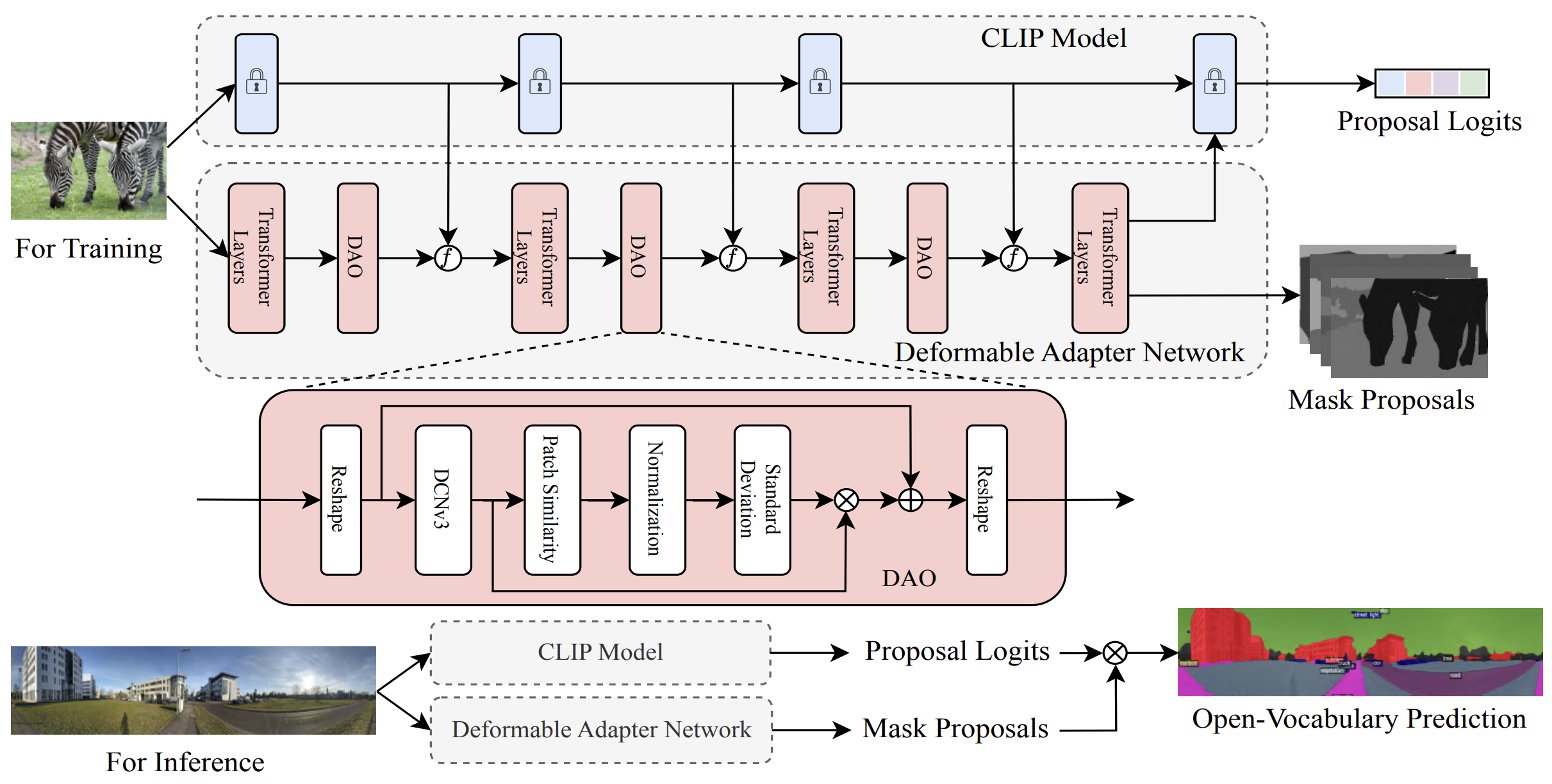}
    \caption{\textbf{Overview of the OOOPS model architecture}. It consists of a frozen CLIP model and a Deformable Adapter Network (DAN) which includes Transformer Layers and the proposed DAO.}
    \label{fig2:model_architecture}
\end{figure*}

\subsection{Model Architecture}
\label{subsec:model_architecture}
A foundation model~\cite{radford2021clip} can be efficiently transferred to downstream tasks by using an adapter~\cite{xu2023san}. To improve the modeling ability for panoramas, we design the OOOPS model.
As shown in Fig.~\ref{fig2:model_architecture}, it comprises a frozen CLIP model and a proposed Deformable Adapter Network (DAN) that incorporates multiple Transformer Layers and the novel DAO, which will be detailed in Sec.~\ref{subsec:dao}. Feature fusion takes place between the intermediate layers of CLIP and DAN. One of the two outputs from DAN consists of mask proposals, while the other serves as deep supervised guidance for CLIP to generate proposal logits. In the training phase, pinhole images are forwarded into OOOPS, generating mask proposals and proposal logits for the loss calculation. In the inference phase, panoramas are forwarded into OOOPS to generate segmentation predictions via the multiplication of mask proposals and the corresponding proposal logits. The frozen CLIP is necessary for the zero-shot learning ability of OOOPS.

\subsection{Deformable Adapter Network}
\label{subsec:dao}
The Deformable Adapter Network is a combination of multiple Transformer~\cite{vaswani2017attention} Layers and the proposed DAO. 
Since distortion exists in panoramic images, which is a big challenge when utilizing informative panoramas, we delve deeper into the deformable design~\cite{wang2023DCNv3} and sampling methods like APES~\cite{wu2023apes}, MateRobot~\cite{zheng2024materobot}, proposing DAO to tackle the image distortions and object deformations of panoramas.
To provide a detailed rethinking process of the deformable convolution design, we introduce the preliminary of the DCN series as follows.

\noindent\textbf{Revisiting DCN Series.} 
The seminal work DCN~\cite{dai2017DCNv1} can enable the traditional CNN with spatial deformation-aware ability.
Given a convolutional kernel of $K$ sampling locations, let $\mathbf{w}_k$ and $\mathbf{p}_k$ denote the weight and pre-specified offset for
the $k$-th location, respectively.
For example, $K{=}9$ and $\mathbf{p}_k {\in} \{(1,1), \ldots, (-1,-1)\}$ defines a $3{\times}3$ convolutional kernel of dilation $1$.
Let $\mathbf{x}(\mathbf{p})$ and $\mathbf{y}(\mathbf{p})$ denote the features at location $\mathbf{p}$ from the input feature maps $\mathbf{x}$ and output feature maps $\mathbf{y}$, respectively.
DCN is formulated as:
\begin{equation}
    \mathbf{y}(\mathbf{p}) = \sum^K_{k=1}\mathbf{w}_k\mathbf{x}(\mathbf{p}+\mathbf{p}_k+\Delta\mathbf{p}_k),
\end{equation}
where $\Delta\mathbf{p}_k$ is the learnable offset for the $k$-th location.
Although DCN is capable of capturing spatial deformation, every sampling location is treated equally when calculating the local features. DCNv2~\cite{zhu2019DCNv2} is proposed with an additional term called modulation scalar. Specifically, DCNv2 can be formulated as:
\begin{equation}
    \mathbf{y}(\mathbf{p}) = \sum^K_{k=1}\mathbf{w}_k\mathbf{m}_k\mathbf{x}(\mathbf{p}+\mathbf{p}_k+\Delta\mathbf{p}_k),
\end{equation}
where $\mathbf{m}_k$ is the learnable modulation scalar for the $k$-th location.
Inspired by Transformer, DCNv3~\cite{wang2023DCNv3} is proposed with a grouping operation, further boosting the deformation-aware ability of DCNv2. DCNv3 can be addressed with the following formula:
\begin{equation}
    \mathbf{y}(\mathbf{p}) = \sum^G_{g=1}\sum^K_{k=1}\mathbf{w}_g\mathbf{m}_{gk}\mathbf{x}_g(\mathbf{p}+\mathbf{p}_k+\Delta\mathbf{p}_{gk}),
\end{equation}
where $G$ denotes the total number of aggregation groups.
DCNv4~\cite{xiong2024DCNv4} is similar to DCNv3, achieving similar performance while significantly reducing the runtime.

\begin{figure*}[h]
\centering
\includegraphics[width=0.5\textwidth]{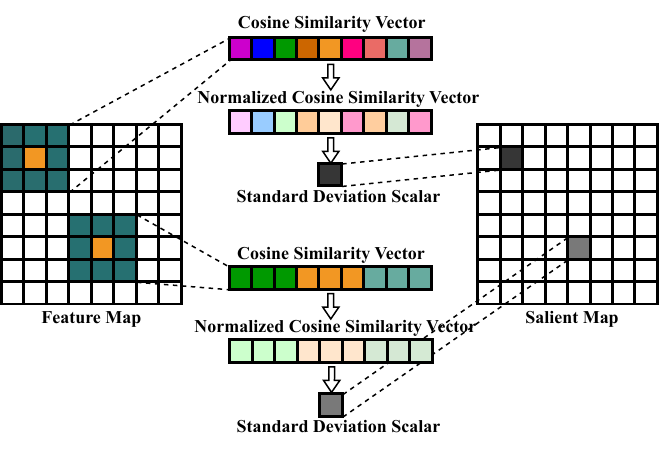}
\caption{Salient map generation in DAO.}
\label{fig3:salient_map}
\end{figure*}
\noindent\textbf{Deformable Adapter Operator (DAO).} When dealing with distortions in panoramas, DCNv3 and DCNv4 fall short of meeting the distortion-aware requirements.
This is illustrated by the visualization of Fig.~\ref{fig6:deformable_offset} in Sec.~\ref{subsec:ablation}.
Therefore, DAO is proposed to tackle the distortion problem in panoramic images using the following expression:
\begin{equation}
    \mathbf{y}(\mathbf{p}) = \mathbf{s}(\mathbf{p})\sum^G_{g=1}\sum^K_{k=1}\mathbf{w}_g\mathbf{m}_{gk}\mathbf{x}_g(\mathbf{p}+\mathbf{p}_k+\Delta\mathbf{p}_{gk}),
\end{equation}
where $\mathbf{s(\mathbf{p}})$ is the learnable salient scalar at location $\mathbf{p}$.
Inheriting from DCNv3, DAO comes up with an additional term called salient scalar indicating the importance of each pixel within the whole panorama.
It's worth noting that DCNv3 and DCNv4 share the same mathematical expression, but DCNv3 is \textbf{more robust} according to our experiments before designing DAO.
Therefore, we adopt DCNv3 as a part of DAO rather than DCNv4.
Illustrated in Fig.~\ref{fig2:model_architecture}, the feature map outputted by DCNv3 passes through the Patch Similarity Layer, Normalization Layer, and Standard Deviation Layer sequentially to form a salient map.
The intuition for such a design is straightforward.
The salient pixels in an image are those that differ significantly from their neighboring pixels, \eg, edge pixels.
If all pixels within an image patch are different, the standard deviation of the pixel similarity for this patch is higher than the one containing similar pixels, resulting in a higher salient scalar.
Fig.~\ref{fig3:salient_map} gives a more detailed explanation of the salient map generation.
Given a feature map, DAO first calculates the cosine similarity between the center pixel and all pixels within a kernel, \eg, $9$ pixels within a $3 \times 3$ kernel in Fig.~\ref{fig3:salient_map}, resulting in a $9$-dimensional cosine similarity vector.
Softmax normalization is then applied to the vector.
Afterward, DAO calculates the standard deviation of this normalized cosine similarity vector, indicating the importance of the center pixel.
By traversing every pixel across the entire feature map, a salient map is generated for the sake of enhancing the salient pixels, which are usually the edge pixels of an image where strong panoramic distortion occurs.

\subsection{Random Equirectangular Projection}
\label{subsec:rerp}
\begin{figure}[t]
    \centering
    \begin{subfigure}[b]{0.45\textwidth}
        \includegraphics[width=0.99\linewidth,height=3.5cm]{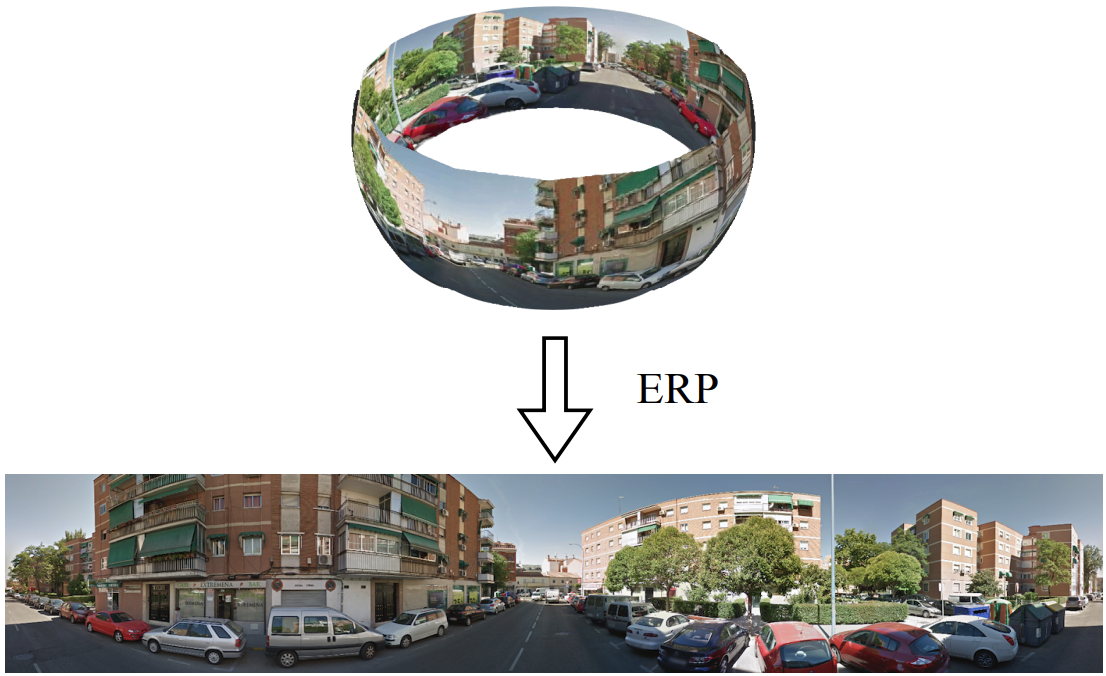}
        \caption{Equirectangular Projection (ERP) maps a globe into a panorama. It's clear that strong distortion occurs in the panorama, especially on object edges in the horizontal direction.}
        \label{fig4-a:erp}
    \end{subfigure}
    \hspace{1mm}
    \begin{subfigure}[b]{0.45\textwidth}
        \includegraphics[width=0.99\linewidth,height=3.5cm]{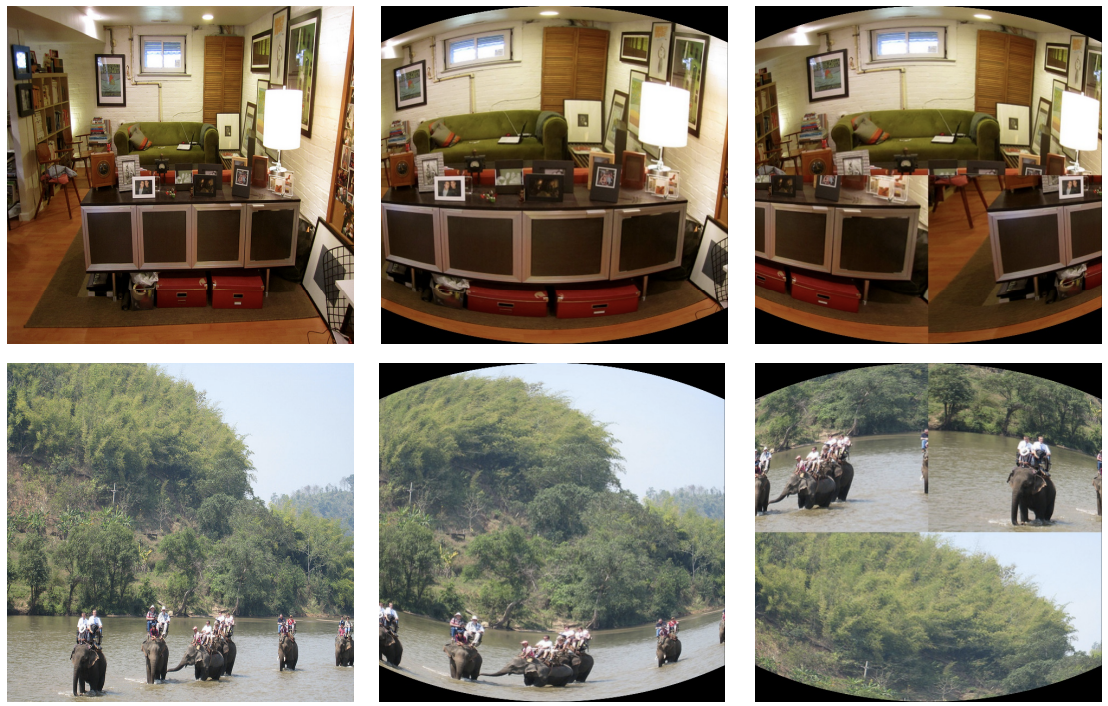}
        \caption{Visualization of ERP and RERP of pinhole images. First column: pinhole images. Second column: ERP on pinhole images. Third column: RERP on pinhole images.}
        \label{fig4-b:rerp}
    \end{subfigure}
    \caption{(a) Visualization of ERP on a panoramic image and (b) Our proposed RERP on pinhole images.}
    \label{fig4:data_augmentation}
\end{figure}
The Equirectangular Projection~\cite{ray2018erp} (ERP) is one of the most common methods for mapping a globe into a panoramic plane, which transforms spherical coordinates into planar coordinates expressed as follows:
\begin{align}
    &x = R(\lambda-\lambda_0)cos(\varphi_1), \\
    &y = R(\varphi-\varphi_0),
\end{align}
where $\lambda$ and $\varphi$ are the longitude and latitude of the location to project, respectively.
$\varphi_1$ are the standard parallels.
$\lambda_0$ and $\varphi_0$ are the central meridian and central parallel of the map, respectively. 
$R$ is the radius of the globe.
$x$ denotes the horizontal coordinate, while $y$ represents the vertical coordinate of the projected location on the map.
Fig.~\ref{fig4-a:erp} visualizes the ERP on a panoramic plane.
It can be observed that after equirectangular projection strong distortion occurs in the panorama, \eg, a straight line is transformed into a curve.
To further boost the performance, we propose Random Equirectangular Projection (RERP) on pinhole images since the OPS task requires the model to train with pinhole images instead of panoramas.

We divide the pinhole image into four parts and randomly shuffle the image patches.
Afterward, we apply equirectangular projection to the distortion-free pinhole image.
Fig.~\ref{fig4-b:rerp} visualizes pinhole images after RERP.
The first column is the pinhole images without any data augmentation.
The second column is ERP on pinhole images without random shuffling.
The last column is our proposed RERP.
It can be observed that the panorama-like distortion also occurs in the pinhole images after RERP.
Random shuffling is employed to enhance robustness and promote generalization.

\section{Experiments}
\label{experiments}

\subsection{Datasets} 
We train our model using the pinhole general-object COCOStuff-164k~\cite{Caesar2016COCOStuff} while evaluating the model performance on three different panoramic datasets: panoramic street-view WildPASS~\cite{yang2021wildpass}, panoramic indoor-scene Stanford2D3D~\cite{armeni1702s2d3d} and panoramic indoor-scene Matterport3D~\cite{Matterport3D}.

\label{subsec:datasets}
\noindent\textbf{COCOStuff-164k.}
The pinhole dataset comprises $164$k images across $171$ annotated classes, partitioned into training, validation, and test sets, with $118$k, $5$k, and $41$k images, respectively. In our experiments, we utilize the entire $118$k-image training set as the training data by default.

\noindent\textbf{WildPASS.}
The panoramic street-view dataset contains $2,500$ panoramas in $65$ cities from all continents Asia, Europe, Africa, Oceania, and North and South America (excluding only Antarctica) with $8$ dense-annotated classes. 
All images have a $70${\textdegree}${\times}$$360${\textdegree} FoV in the size of $400{\times}2048$.

\noindent\textbf{Stanford2D3D.}
The panoramic indoor dataset consists of $1,413$ images in the size of $512{\times}1024$ and $13$ object classes. 
All results are averaged over $3$ cross-validation folds.

\noindent\textbf{Matterport3D.}
The panoramic indoor dataset includes $10,800$ panoramic views from $194,400$ RGB-D images.
According to the data preprocessing pipeline of 360BEV~\cite{teng2023360bev}, a subset of $772$ images in the size of $512{\times}1024$ with $20$ classes is used to evaluate the segmentation performance.

\subsection{Metrics}
\label{subsec:metrics}
\noindent\textbf{Close-vocabulary Metric.} Mean Intersection over Union (mIoU) is used as the evaluation metric in close-vocabulary semantic segmentation tasks. 
Specifically, IoU is calculated as the area of overlap between the predicted segmentation and the ground truth, divided by the area of union between them.
mIoU is a hard metric designed for close-vocabulary segmentation ignoring the semantic meanings of similar classes referring to the open-vocabulary setting.

\noindent\textbf{Open-vocabulary Metric.} Open mean Intersection over Union~\cite{zhou2023rethinking_open}, open mIoU for short, is a soft metric compared to mIoU since the similarity scores between different classes are also taken into account, which are calculated by the WordNet approach~\cite{miller1995wordnet}.
In this work, we will report both close and open metrics for a comprehensive understanding of all models.

\subsection{Implementation Details}
\label{subsec:implementation_details}
We utilize CLIP ViT-B/$16$ as the frozen CLIP model. All models are trained with the pinhole general-object COCOStuff-164k training set with $4\times$A$40$ GPUs while evaluated on three panoramic datasets which are in totally different domains compared with the pinhole dataset.
The AdamW optimizer is applied with an initial learning rate of $0.0001$, weight decay of $0.0001$, batch size of $32$, and a total of $60,000$ training iterations.
Throughout the training, the learning rate follows a polynomial schedule with a power of $0.9$.
Data augmentations, including random image resizing within the short-side range of $[320,1024]$ and a crop size of $640{\times}640$ are applied.
Additionally, when verifying the effectiveness of the proposed RERP, it is applied as an extra data augmentation.
Following the practice of SAN~\cite{xu2023san}, apart from the supervision of mask recognition with cross-entropy loss, we utilize the dice loss and binary cross-entropy loss for mask proposal generation.

\subsection{Comparison with the State of the Art}
\label{subsec:comparison_with_sota}
\noindent\textbf{Results on WildPASS.}
\begin{table*}[t]
\tiny
\setlength{\abovecaptionskip}{0pt}
\setlength{\belowcaptionskip}{0pt}
\caption{\textbf{Results on WildPASS}. The mIoU, open mIoU, and per-class IoU are in percentage (\%). \#PARAMs is the number of learnable parameters in millions.}
\label{tab:wildpass_sota}
\begin{center}
\resizebox{\linewidth}{!}{
\setlength{\tabcolsep}{1pt}
\renewcommand{\arraystretch}{0.99}
\begin{tabular}{lccccccccccc}
\toprule[1pt]
{\textbf{Network}}&{\textbf{\rotatebox{45}{\#PARAMs}}}&{\textbf{\rotatebox{45}{mIoU}}}&{\textbf{\rotatebox{45}{Open mIoU}}}&{\textbf{\rotatebox{45}{Car}}}&{\textbf{\rotatebox{45}{Road}}}&{\textbf{\rotatebox{45}{Sidewalk}}}&{\textbf{\rotatebox{45}{Crosswalk}}}&{\textbf{\rotatebox{45}{Curb}}}&{\textbf{\rotatebox{45}{Person}}}&{\textbf{\rotatebox{45}{Truck}}}&{\textbf{\rotatebox{45}{Bus}}}\\
\midrule\midrule
\rowcolor{blue!15}\multicolumn{12}{c}{Close-Vocabulary}\\
\midrule
{Fast-SCNN~\cite{poudel2019fast}}&\textbf{1.1}&{24.8}&-&{45.9}&{60.0}&{31.7}&{9.7}&{17.1}&{6.0}&{14.2}&{13.8}\\
{PSPNet18~\cite{zhao2017pyramid}}&{17.5}&{28.2}&-&{58.2}&{66.8}&{28.4}&{13.0}&{19.2}&{6.2}&{18.2}&{15.6}\\
{SwiftNet~\cite{orvsic2019defense}}&{11.8}&{30.0}&-&{55.7}&{64.1}&{29.2}&{16.2}&{22.9}&{8.5}&{21.1}&{22.2}\\
{DenseASPP~\cite{yang2018denseaspp}}&{8.3}&{33.2}&-&{51.1}&{69.1}&{38.1}&{16.4}&{26.3}&{8.7}&{27.4}&{28.4}\\
{ERF-PSPNet~\cite{yang2020pass}}&{2.5}&{34.0}&-&{66.3}&{70.5}&{36.5}&{6.4}&{24.1}&{9.4}&{26.5}&{32.0}\\
{PSPNet50~\cite{zhao2017pyramid}}&{53.3}&{46.1}&-&{80.0}&{74.9}&{51.7}&{23.9}&{31.4}&{19.8}&{38.9}&{48.1}\\
{DANet~\cite{fu2019dual}}&{47.4}&{47.2}&-&{74.8}&{72.2}&{49.9}&{28.9}&{23.8}&{25.4}&{51.9}&{50.6}\\
{OOSS (ERF-PSPNet)~\cite{yang2020omnisupervised}}&{2.5}&{56.1}&-&{87.2}&{79.3}&{60.8}&{28.0}&{38.1}&{54.5}&{48.8}&{52.2}\\
{PASS (ERF-PSPNet)~\cite{yang2020pass}}&{2.5}&{64.7}&-&{87.3}&{80.0}&{61.4}&{{71.1}}&{49.9}&{\textbf{72.2}}&{37.5}&{57.9}\\
{ERF-PSPNet (omni-sup')~\cite{yang2021wildpass}}&{2.5}&{66.8}&-&{90.5}&{82.7}&{65.6}&{70.5}&{51.5}&{58.2}&{\textbf{62.0}}&{53.1}\\
{ECANet (with attention)~\cite{yang2021wildpass}}&{2.6}&{67.7}&-&{90.4}&{83.7}&{\textbf{68.4}}&{67.8}&{\textbf{52.1}}&{61.4}&{54.5}&{\textbf{63.3}}\\
{ECANet (with fusion)~\cite{yang2021wildpass}}&{2.6}&{\textbf{69.0}}&-&{\textbf{90.6}}&{\textbf{85.7}}&{68.0}&{\textbf{67.9}}&{\textbf{52.1}}&{66.0}&{59.3}&{62.3}\\ 
\midrule
\rowcolor{blue!15}\multicolumn{12}{c}{Open-Vocabulary}\\
\midrule
{SAN~\cite{xu2023san}}& \textbf{8.4} & 55.6 & 56.9 & 84.3 & \textbf{81.9} & 69.6 & 10.1 & 2.8 & 72.2 & 54.9 & 69.1 \\
{CAT-seg~\cite{cho2023cat_seg}}& 59.5 & 55.7 & 57.1 & 84.1 & 81.5 & \textbf{70.0} & 10.2 & 3.0 & 72.5 & 54.8 & 69.3 \\
{OpenSeeD~\cite{zhang2023openseed}}& 65.4 & 55.8 & 57.2 & 84.3 & 81.6 & 69.7 & 10.3 & 3.0 & 73.6 & 54.9 & 69.1 \\
\rowcolor{gray!10}{OOOPS (ours, w/o RERP)}& 8.7 & 57.0 & 58.5 & \textbf{84.9} & 81.6 & 68.7 & 15.1 & \textbf{8.4} & 71.3 & \textbf{55.3} & 70.4 \\
\rowcolor{gray!10}{OOOPS (ours, w/ RERP)}& 8.7 &\textbf{58.0} & \textbf{59.7} & \textbf{84.9} & 81.6 & 68.7 & \textbf{19.4} & 8.3 & \textbf{73.8} & 54.9 & \textbf{72.2}\\
\bottomrule
\end{tabular}
}
\end{center}
\end{table*}

\begin{figure*}[t]
    \centering
    \begin{subfigure}[b]{0.49\textwidth}
        \centering
        \includegraphics[width=1.0\columnwidth,height=2.5cm]{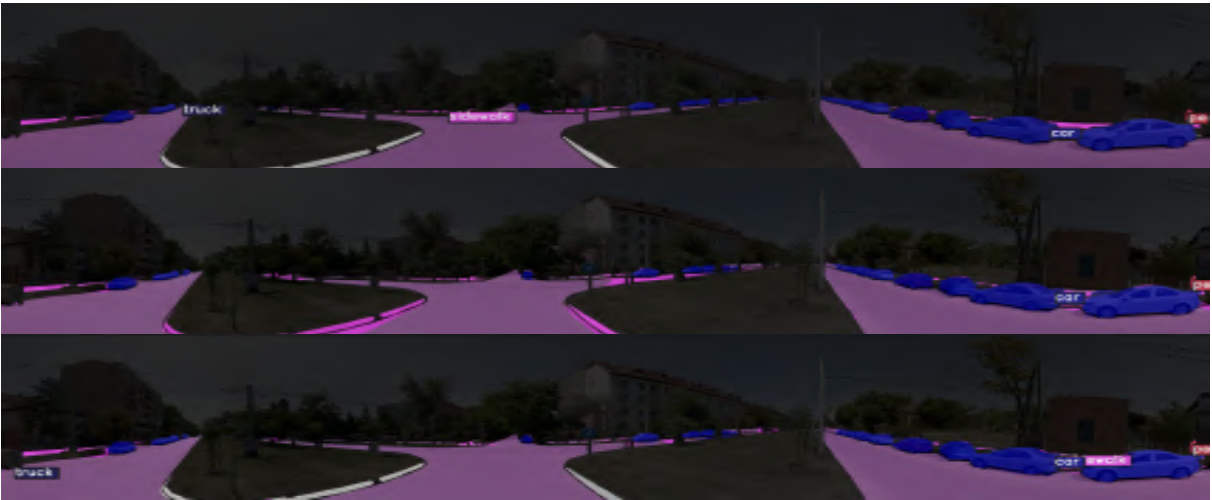}
        \caption{Comparison with baseline method SAN~\cite{xu2023san} on the WildPASS dataset with predefined $8$ classes. First row: ground truth. Second row: prediction from SAN. Third row: prediction from OOOPS with RERP.}
        \label{fig5-a:wildpass_close_vocabulary_visualization}
    \end{subfigure}
    \hspace{0.3mm}
    \begin{subfigure}[b]{0.49\textwidth}
        \centering
        \includegraphics[width=1.0\columnwidth,height=2.5cm]{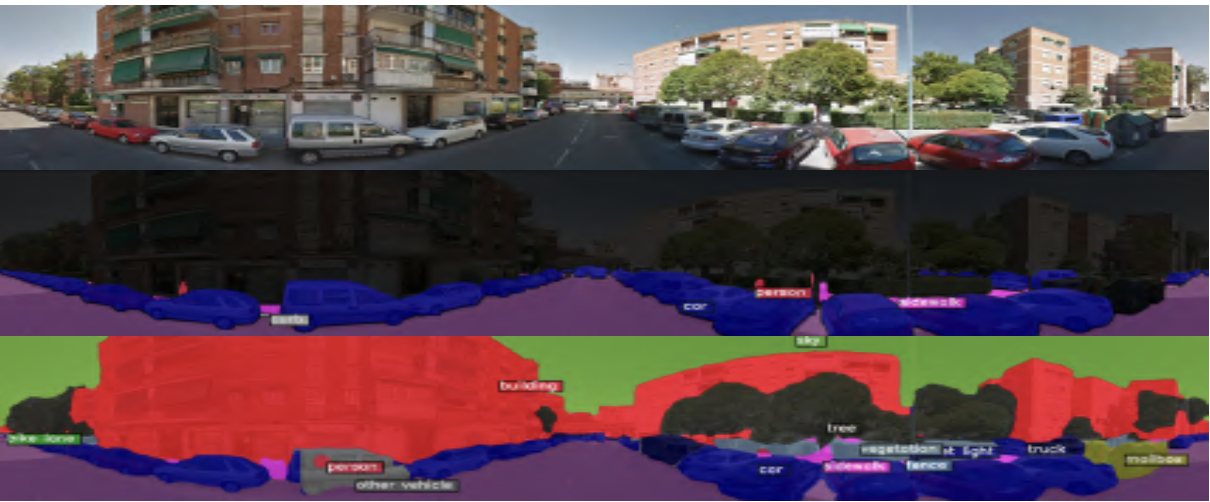}
        \caption{Visualization of the prediction from OOOPS in close- and open-vocabulary settings. First row: RGB. Second row: prediction with close vocabularies. Third row: prediction with open vocabularies. Zoom in for a better view.}
        \label{fig5-b:OPS_prediction_for_close_and_open_vocabularies}
    \end{subfigure}
    \caption{(a) Comparison on the WildPASS dataset and (b) Visualization of the prediction from OOOPS in close- and open-vocabulary settings.}
    \label{fig5:comparison_with_sota}
\end{figure*}
Comparisons with both close- and open-vocabulary state-of-the-art methods on WildPASS are listed in Table~\ref{tab:wildpass_sota}.
It is not surprising that ECANet (with fusion)~\cite{yang2021wildpass} achieves the best performance among all methods since it is a close-vocabulary method, meaning that ECANet is only able to predict the classes predefined by the dataset.
However, our OOOPS model is qualified for predicting classes that are not annotated in the dataset.
The OOOPS method (with RERP) surpasses other state-of-the-art open-vocabulary panoramic segmentation methods with over $\textbf{{+}2.2}\%$ mIoU while introducing only a minor increment of $0.3$ million learnable parameters compared to the lightweight baseline SAN~\cite{xu2023san}, offering a promising step towards bridging the gap between close- and open-vocabulary settings.
Moreover, our OOOPS method enhanced by RERP demonstrates significant advancements in challenging categories such as \textit{crosswalk} and \textit{curb}, achieving $\textbf{{+}9.3}\%$ and $\textbf{{+}5.5}\%$ mIoU compared with the baseline SAN, respectively.
It is worth noting that in some categories, \eg, \textit{sidewalk}, \textit{person}, and \textit{bus}, our proposed OOOPS (both with and without RERP) even outperforms the models trained in a close-vocabulary setting.
All open-vocabulary methods achieve better performance with open mIoU metric compared to the close mIoU since similar semantic meanings are also considered in the calculation.
Fig.~\ref{fig1-a:performance_drop} showcases the performance degradation on both metrics.
Even though all methods have degraded performance when FoV increases, our OOOPS achieves a smaller drop (${-}\textbf{10.0}\%$) while others have a larger performance decrease (${-}12.0\%$), indicating the effectiveness of the OOOPS model.

Fig.~\ref{fig5-a:wildpass_close_vocabulary_visualization} visualizes the prediction of the challenging \textit{curb} category from baseline SAN and our proposed OOOPS with RERP.
From top to bottom are the ground truth, prediction from SAN, and prediction from OOOPS with RERP, respectively.
The baseline SAN cannot recognize the challenging \textit{curb} category while our proposed OOOPS with RERP is capable of detecting the \textit{curb} in the middle of the panorama highlighted in gray.
Fig.~\ref{fig5-b:OPS_prediction_for_close_and_open_vocabularies} showcases the visualization of the prediction from OOOPS on WildPASS in close- and open-vocabulary settings. From top to bottom are input RGB, prediction from OOOPS with $8$ predefined close vocabularies from WildPASS, and prediction from OOOPS with unlimited open vocabularies.
It is evident that our OOOPS with RERP is capable of tackling unseen categories in the open-vocabulary setting.
Every pixel on the panorama is assigned a meaningful semantic label.
For example, ``mailbox'' is not a predefined category from WildPASS.
While existing models can not recognize the \textit{mailbox} under a close-vocabulary setting, the rare classes can be easily detected by our model in an open-vocabulary one, indicating the great potential of open-vocabulary panoramic segmentation for parsing unconstrained traffic surroundings.

\begin{table*}[t]
\scriptsize
\setlength{\abovecaptionskip}{0pt}
\setlength{\belowcaptionskip}{0pt}
\caption{\textbf{Results on Stanford2D3D}. The mIoU, open mIoU, and per-class IoU are in percentage (\%). \#PARAMs is the number of learnable parameters in millions.}
\label{tab:s2d3d_sota}
\begin{center}
\resizebox{\linewidth}{!}{
\renewcommand{\arraystretch}{0.99}
\begin{tabular}{lcccccccccccccccccccccc}
\toprule[1pt]
{\textbf{Network}}&{\textbf{\rotatebox{65}{\#PARAMs}}}&{\textbf{\rotatebox{65}{mIoU}}}&{\textbf{\rotatebox{65}{Open mIoU}}}&{\textbf{\rotatebox{65}{Beam}}}&{\textbf{\rotatebox{65}{Board}}}&{\textbf{\rotatebox{65}{Bookcase}}}&{\textbf{\rotatebox{65}{Ceiling}}}&{\textbf{\rotatebox{65}{Chair}}}&{\textbf{\rotatebox{65}{Clutter}}}&{\textbf{\rotatebox{65}{Column}}}&{\textbf{\rotatebox{65}{Door}}}&{\textbf{\rotatebox{65}{Floor}}}&{\textbf{\rotatebox{65}{Sofa}}}&{\textbf{\rotatebox{65}{Table}}}&{\textbf{\rotatebox{65}{Wall}}}&{\textbf{\rotatebox{65}{Window}}}\\
\midrule\midrule
\rowcolor{blue!15}\multicolumn{17}{c}{Close-Vocabulary}\\
\midrule
Trans4PASS~\cite{zhang2022trans4pass} & \textbf{28.2} & 52.1 & - & \textbf{11.4} & 60.6 & 51.1 & 81.8 & 59.7 & 37.9 & \textbf{23.5} & {54.6} & 88.8 & {34.6} & 53.9 & 68.1 & 51.0 \\
360BEV\cite{teng2023360bev} & 28.4 &\textbf{54.3}& - & 06.0 & \textbf{63.6}&	\textbf{55.5}&	\textbf{82.7}&	\textbf{62.7}&	\textbf{39.9}&	16.4&	\textbf{58.8}&	\textbf{90.8}&	\textbf{45.2}&	\textbf{57.1}&	\textbf{71.1}&	\textbf{57.6}\\
\midrule
\rowcolor{blue!15}\multicolumn{17}{c}{Open-Vocabulary}\\
\midrule
{SAN~\cite{xu2023san}}& \textbf{8.4} & 38.5 & 39.5 & \textbf{0.2} & 51.5 & 39.6 & 66.5 & 52.1 & \textbf{10.7} & 0.0 & 45.0 & 49.3 & 31.8 & \textbf{49.2} & 50.0 & 53.4 \\
{CAT-seg~\cite{cho2023cat_seg}}& 59.5 & 38.6 & 39.6 & 0.0 & 51.6 & 42.3 & 66.8 & 51.4 & 8.0 & 0.0 & 47.4 & 50.7 & 31.2 & 47.5 & 51.4 & 53.5 \\
{OpenSeeD~\cite{zhang2023openseed}}& 65.4 & 38.7 & 40.0 & 0.0 & 51.8 & 42.0 & 67.0 & 51.6 & 8.4 & \textbf{0.2} & 47.5 & 50.0 & 31.4 & 47.1 & 51.7 & 53.6 \\
\rowcolor{gray!10}{OOOPS (w/o RERP)}& 8.7 & 39.5 & 41.0 & 0.0 & \textbf{52.3} & 44.1 & 67.1 & \textbf{53.7} & 9.7 & 0.0 & \textbf{50.1} & 52.8 & 30.5 & 47.2 & 52.2 & 53.8 \\
\rowcolor{gray!10}{OOOPS (w/ RERP)}& 8.7 & \textbf{41.1} & \textbf{42.6} & 0.0 & 50.7 & \textbf{44.8} & \textbf{68.8} & 51.8 & 8.1 & 0.0 & 48.2 & \textbf{71.1} & \textbf{33.6} & 47.2 & \textbf{55.4} & \textbf{54.5} \\ 
\bottomrule
\end{tabular}
}
\end{center}
\end{table*}
\noindent\textbf{Results on Stanford2D3D.}
In Table~\ref{tab:s2d3d_sota}, the comparison is conducted between state-of-the-art methods on the Stanford2D3D dataset.
All reported results are averaged over $3$ cross-validation folds.
Without RERP, OOOPS achieves $\textbf{39.5\%}$ mIoU, outperforming other open-vocabulary methods like SAN~\cite{xu2023san}, CAT-seg~\cite{cho2023cat_seg}, and OpenSeeD~\cite{zhang2023openseed}, which score between $38.5\%$ and $38.7\%$. With RERP, OOOPS's mIoU increases to $\textbf{41.1\%}$ with a $\textbf{{+}2.4\%}$ gain, narrowing the gap with advanced close-vocabulary methods.

\begin{table*}[t]
\scriptsize
\setlength{\abovecaptionskip}{0pt}
\setlength{\belowcaptionskip}{0pt}
\caption{\textbf{Results on Matterport3D}. The mIoU, open mIoU, and per-class IoU are in percentage (\%). \#PARAMs is the number of learnable parameters in millions.}
\label{tab:mp3d_sota}
\begin{center}
\resizebox{\linewidth}{!}{
\renewcommand{\arraystretch}{0.99}
\begin{tabular}{lccccccccccccccccccccccc}
\toprule[1pt]
{\textbf{Network}}&{\textbf{\rotatebox{65}{\#PARAMs}}}&{\textbf{\rotatebox{65}{mIoU}}}&{\textbf{\rotatebox{65}{Open mIoU}}}&{\textbf{\rotatebox{65}{Wall}}}&{\textbf{\rotatebox{65}{Floor}}}&{\textbf{\rotatebox{65}{Chair}}}&{\textbf{\rotatebox{65}{Door}}}&{\textbf{\rotatebox{65}{Table}}}&{\textbf{\rotatebox{65}{Picture}}}&{\textbf{\rotatebox{65}{Furniture}}}&{\textbf{\rotatebox{65}{Objects}}}&{\textbf{\rotatebox{65}{Window}}}&{\textbf{\rotatebox{65}{Sofa}}}&{\textbf{\rotatebox{65}{Bed}}}&{\textbf{\rotatebox{65}{Sink}}}&{\textbf{\rotatebox{65}{Stairs}}}&{\textbf{\rotatebox{65}{Ceiling}}}&{\textbf{\rotatebox{65}{Toilet}}}&{\textbf{\rotatebox{65}{Mirror}}}&{\textbf{\rotatebox{65}{Shower}}}&{\textbf{\rotatebox{65}{Bathtub}}}&{\textbf{\rotatebox{65}{Counter}}}&{\textbf{\rotatebox{65}{Shelving}}}\\
\midrule\midrule
\rowcolor{blue!15}\multicolumn{24}{c}{Close-Vocabulary}\\
\midrule
Trans4PASS+ \cite{trans4passplus} & 28.5 & 42.6 & - & 63.4 & 79.1 & 39.1 & 40.3 & 32.8 & 36.0 & 31.0 & 31.5 & 37.5 & 44.0 & 63.2 & 20.6 & 41.8 & 77.6 & 40.7 & 24.3 & \textbf{23.7} & 58.3 & 34.3 & 32.9 \\
360BEV~\cite{teng2023360bev} & \textbf{28.4} &\textbf{46.4} & - & \textbf{64.1} & \textbf{83.1} & \textbf{45.8} & \textbf{45.0} & \textbf{38.0} & \textbf{41.1} & \textbf{32.3} & \textbf{35.1} & \textbf{40.6} & \textbf{48.7} & \textbf{69.8} & \textbf{25.1} & \textbf{47.8} & \textbf{80.2} & \textbf{46.0} & \textbf{28.7} & 22.3 & \textbf{60.1} & \textbf{38.6} & \textbf{34.8} \\
\midrule
\rowcolor{blue!15}\multicolumn{24}{c}{Open-Vocabulary}\\
\midrule
{SAN~\cite{xu2023san}}& \textbf{8.4} & 30.5 & 31.4 & 54.4 & 76.1 & 18.0 & 34.1 & 22.3 & 21.1 & 13.0 & 1.7 & 32.7 & 34.2 & 51.0 & 11.0 & 19.9 & 74.2 & 23.4 & 12.0 & 31.6 & 31.9 & 19.0 & 28.4 \\
{CAT-seg~\cite{cho2023cat_seg}}& 59.5 & 30.2 & 31.1 & 54.3 & 74.8 & 17.0 & 34.1 & 21.9 & 20.8 & 12.6 & 1.9 & 32.9 & 34.0 & 50.1 & 10.7 & 18.8 & 73.5 & 22.4 & 11.6 & 31.9 & 32.9 & 19.2 & 28.5 \\
{OpenSeeD~\cite{zhang2023openseed}}& 65.4 & 30.6 & 31.6 & 54.2 & 76.3 & 18.1 & 34.3 & 22.5 & 21.0 & 13.0 & 2.1 & 32.8 & 34.1 & 50.9 & 11.2 & 20.2 & 74.1 & 23.5 & 12.2 & 31.1 & 31.8 & 19.1 & 28.6 \\
\rowcolor{gray!10}{OOOPS (w/o RERP)}& 8.7 & 31.1 & 32.4 & 55.2 & \textbf{77.0} & 17.1 & \textbf{35.7} & \textbf{25.0} & 17.5 & \textbf{13.5} & 2.3 & 33.0 & 32.4 & 51.8 & \textbf{12.6} & \textbf{20.4} & 74.5 & 22.6 & 10.8 & \textbf{34.1} & \textbf{36.8} & \textbf{19.4} & \textbf{29.7} \\
\rowcolor{gray!10}{OOOPS (w/ RERP)}& 8.7 & \textbf{31.2} & \textbf{32.5} & \textbf{55.4} & 76.4 & \textbf{18.4} & 34.8 & 23.9 & \textbf{21.2} & 13.2 & \textbf{2.6} & \textbf{33.3} & \textbf{34.5} & \textbf{52.9} & 12.1 & 20.3 & \textbf{74.6} & \textbf{28.2} & \textbf{12.5} & 31.9 & 29.3 & \textbf{19.4} & 28.6 \\ 
\bottomrule
\end{tabular}
}
\end{center}
\end{table*}
\noindent\textbf{Results on Matterport3D.}
In Table~\ref{tab:mp3d_sota}, the comparison is conducted between state-of-the-art methods on the Matterport3D dataset.
Our OOOPS method showcases consistent performance. 
Without augmentation, OOOPS achieves mIoU of $\textbf{31.1\%}$, slightly outperforming other open-vocabulary methods like SAN~\cite{xu2023san} ($30.5\%$), CAT-seg~\cite{cho2023cat_seg} ($30.2\%$), and OpenSeeD~\cite{zhang2023openseed} ($30.6\%$). 
With augmentation, OOOPS's mIoU marginally increases to $\textbf{31.2\%}$, further reinforcing its lead over these open-vocabulary methods.
Both variants of OOOPS still lag behind 
close-vocabulary methods like Trans4PASS+~\cite{trans4passplus} and 360BEV~\cite{teng2023360bev}, which however, are trained in a supervised fashion in a close-vocabulary setting on panoramas and have mIoUs ranging from $42.6\%$ to $46.4\%$.

\begin{table}[h]
\centering
\caption{Task difference between Domain Adaptation (DA) and the proposed OPS.
}
\label{tab:task_difference}
\setlength{\tabcolsep}{6mm}
\resizebox{\columnwidth}{!}{
\renewcommand{\arraystretch}{1}
\begin{tabular}{cccc!{\vrule width 1pt}ccc}
\toprule[2pt]
\multirow{2}{*}{\textbf{Task}} & \multicolumn{2}{c}{\textbf{Domain}} & \multirow{2}{*}{\textbf{\#Classes}}& \multicolumn{3}{c}{\textbf{mIoU (\%)}} \\
\cmidrule[1pt]{2-3} \cmidrule[1pt]{5-7} 
~ & \textbf{Source} & \textbf{Target} & ~ & \textbf{WildPASS} & \textbf{Matterport3D} & \textbf{Stanford2D3D} \\
\midrule[1pt]
DA & \cmark & \cmark & limited & \textbf{72.2} & \textbf{47.2} & \textbf{55.4} \\
OPS & \cmark & \xmark & unlimited & 58.0 & 31.2 & 41.1 \\
\bottomrule[2pt]
\end{tabular}
}
\end{table}
\noindent\textbf{Results in DA Subtask.}
The proposed OPS task differs from Domain Adaptation since no data from the target domain is available to models in OPS while models have access to the data in DA during training.
Moreover, models are required to recognize an unlimited amount of classes in OPS while only a limited number of classes can be recognized in DA.
DA is a special case of OPS, meaning that models in OPS can be easily adapted to DA and achieve better performance.
The converse doesn't hold.
Table~\ref{tab:task_difference} showcases the task difference between OPS and DA.
It can be observed that over ${+}14\%$ gain in mIoU is achieved when adapting our OOOPS model in the DA subtask.

\subsection{Ablation Study}
\label{subsec:ablation}
\begin{table}[!t]
    \footnotesize
    \centering
    \caption{\textbf{Ablation study} of OOOPS components and the number of shuffling patches.}
    \label{tab:ablation}
    \begin{minipage}{0.49\linewidth}
        \centering
        \subcaption{\textbf{Ablation study of OOOPS components on the WildPASS Dataset.} The mIoU is reported in percentage (\%). \#PARAMs stands for the number of learnable parameters in millions.}
        \label{tab:ablation_components}
        \renewcommand{\arraystretch}{0.7}
        \setlength{\tabcolsep}{14pt}
        \scalebox{0.43}{
        \begin{tabular}{l c c c l}
            \toprule[1.5pt]
            \textbf{Method} & \textbf{ERP} & \textbf{RERP} & \textbf{\#PARAMs} & \textbf{mIoU} \\ \midrule\midrule
            SAN~\cite{xu2023san} & \xmark & \xmark & \textbf{8.4} & 55.6 \\
            SAN~\cite{xu2023san} & \cmark & \xmark & \textbf{8.4} & 56.0 \\
            \rowcolor{gray!10}SAN~\cite{xu2023san} & \xmark & \cmark & \textbf{8.4} & 56.5~\gbf{(+0.9)} \\
            \midrule
            + DCN & \xmark & \xmark & 8.6 & 55.6 \\
            + DCN & \cmark & \xmark & 8.6 & 56.0 \\
            \rowcolor{gray!10}+ DCN & \xmark & \cmark & 8.6 & 56.6~\gbf{(+1.0)} \\
            \midrule
            + DCNv2 & \xmark & \xmark & 8.7 & 55.7 \\
            + DCNv2 & \cmark & \xmark & 8.7 & 56.4 \\
            \rowcolor{gray!10}+ DCNv2 & \xmark & \cmark & 8.7 & 56.6~\gbf{(+0.9)} \\
            \midrule
            + DCNv3 & \xmark & \xmark & 8.7 & 55.8 \\
            + DCNv3 & \cmark & \xmark & 8.7 & 56.4 \\
            \rowcolor{gray!10}+ DCNv3 & \xmark & \cmark & 8.7 & 56.8~\gbf{(+1.0)} \\
            \midrule
            + DCNv4 & \xmark & \xmark & 8.7 & 55.8 \\
            + DCNv4 & \cmark & \xmark & 8.7 & 56.4 \\
            \rowcolor{gray!10}+ DCNv4 & \xmark & \cmark & 8.7 & 56.8~\gbf{(+1.0)} \\
            \midrule
            + DAO (ours) & \xmark & \xmark & 8.7 & 57.0 \\
            + DAO (ours) & \cmark & \xmark & 8.7 & 57.6 \\
            \rowcolor{gray!10}+ DAO (ours) & \xmark & \cmark & 8.7 & \textbf{58.0}~\gbf{(+1.0)} \\
            \bottomrule[1.5pt]
        \end{tabular}}
    \end{minipage}
    \hfill
    \begin{minipage}{0.49\linewidth}
        \centering
        \renewcommand{\arraystretch}{1.45}
        \subcaption{\textbf{Ablation study of different numbers of shuffling patches on WildPASS, Matterport3D and Stanford2D3D datasets.} The mIoU is reported in percentage (\%).}
        \label{tab:ablation_num_patches}
        \scalebox{0.8}{
        \begin{tabular}{l | c c c c}
            \toprule[1pt]
            \diagbox{\textbf{Dataset}}{\textbf{mIoU (\%)}}{\textbf{\#Patches}} & 4 & 9 & 16 & 25 \\ \midrule \midrule
            WildPASS & 58.0 & \textbf{58.5} & 58.2 & 58.0\\ \midrule
            Matterport3D & 31.2 & \textbf{31.6} & 31.4 & 31.0\\ \midrule
            Stanford2D3D & 41.1 & \textbf{41.9} & 41.5 & 41.1\\ 
            \bottomrule[1pt]
        \end{tabular}}
    \end{minipage}
\end{table}

We replace DAO in OOOPS, as illustrated in Fig.~\ref{fig2:model_architecture}, with DCN~\cite{dai2017DCNv1}, DCNv2~\cite{zhu2019DCNv2}, DCNv3~\cite{wang2023DCNv3} and DCNv4~\cite{xiong2024DCNv4}.
When referring to the component substitution without data augmentation in Table~\ref{tab:ablation_components},  it is clear that our DAO surpasses other counterparts with a maximum performance gain of $\textbf{{+}1.4\%}$ mIoU.
As for the component substitution with ERP, all models achieve a boost in the range of ${+}0.3\%$ to ${+}0.7\%$ compared with the one without data augmentation.
The gain is even bigger considering the component substitution with RERP, achieving ${+}1.0\%$ mIoU.
It is worth noting that RERP is able to boost the model performance stably among all models.
The best model performance is achieved by the combination of DAO and RERP, yielding $\textbf{58.0\%}$ mIoU with only a slight increment in parameters.
Table~\ref{tab:ablation_num_patches} showcases the influence of different numbers of shuffling patches. The performance is not proportional to the number of patches, indicating the importance of moderate content continuity during training.

\begin{figure*}[t]
    \centering
    \begin{subfigure}[b]{0.49\textwidth}
        \centering
        \includegraphics[width=1.0\columnwidth,height=4cm]{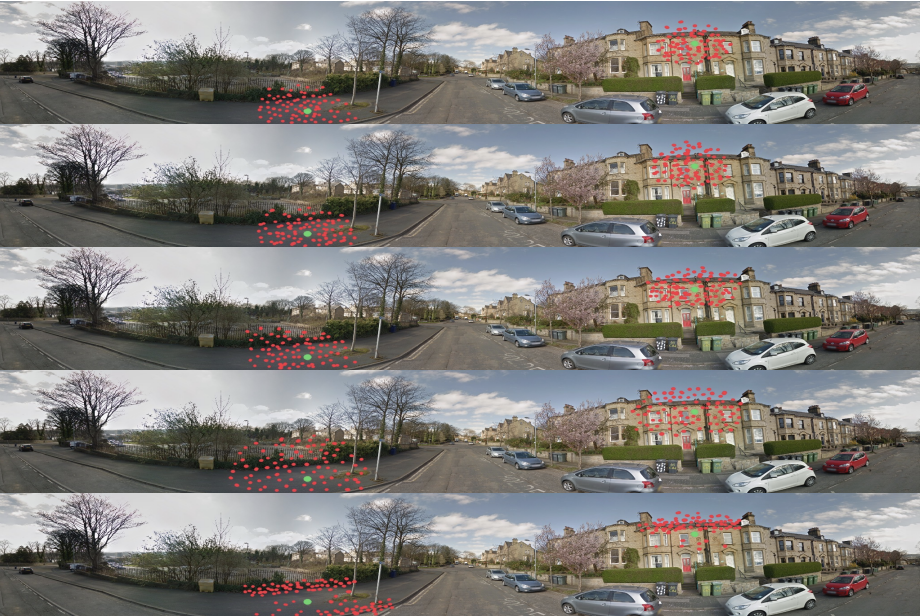}
        \caption{Deformable offsets \textbf{without} RERP. From top to bottom are deformable offsets of DCN, DCNv2, DCNv3, DCNv4, and proposed DAO.}
        \label{fig6-a:dcn_offset_vis}
    \end{subfigure}
    \hspace{0.1mm}
    \begin{subfigure}[b]{0.49\textwidth}
        \centering
        \includegraphics[width=1.0\columnwidth,height=4cm]{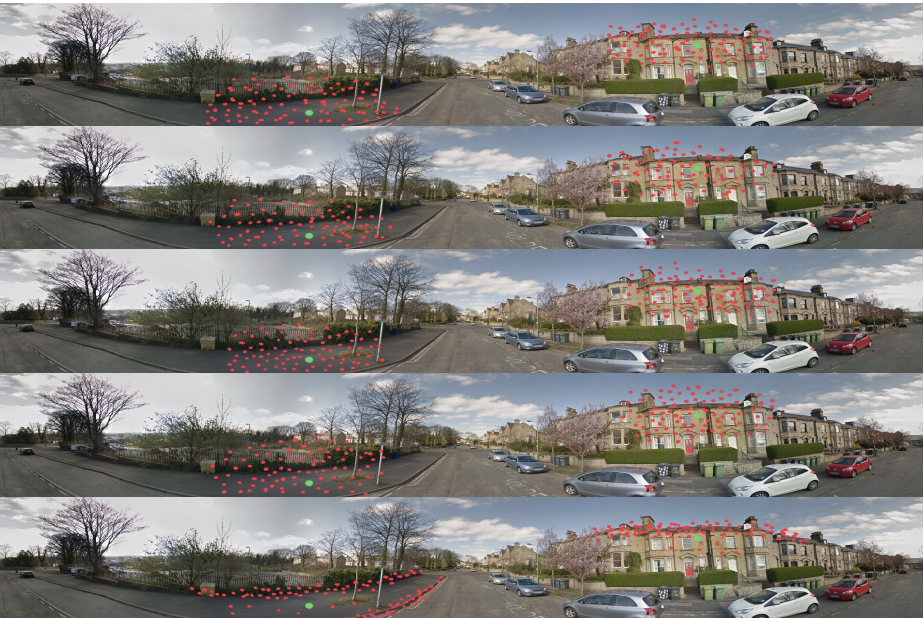}
        \caption{Deformable offsets \textbf{with} RERP. From top to bottom are deformable offsets of DCN, DCNv2, DCNv3, DCNv4, and proposed DAO.}
        \label{fig6-b:dcn_rerp_offset_vis}
    \end{subfigure}
    \caption{\textbf{Visualization of deformable offsets.} The \textcolor{Green}{green points} \textcolor{Green}{$\bullet$} are sampling locations. The \textcolor{red}{red points} \textcolor{red}{$\bullet$} are deformable offsets in $2$ levels, indicating a deformable receptive field (\eg, each level has a $3{\times3}$ kernel size, resulting in $(3{\times}3)^2{=}81$ red points).}
    \label{fig6:deformable_offset}
\end{figure*}

To further illustrate the deformable ability of DAO and the superiority of RERP when tackling the distortion of panoramas, we visualize the deformable offset in Fig.~\ref{fig6:deformable_offset}.
Fig.~\ref{fig6-a:dcn_offset_vis} presents the deformable offsets without RERP.
It can be observed that DAO is better aware of the panoramic distortion compared with the rest in the DCN series.
Since the strongest distortion occurs on the edge of objects, DAO focuses more on edge pixels for better distortion-aware feature extraction.
The standard deviation of patch similarity in DAO implies the salience of the image patch center.
By multiplicatively combining salient scalars and features, significant pixels, such as those representing object edges, are enhanced in their representation.
In Fig.~\ref{fig6-b:dcn_rerp_offset_vis}, it is evident that the inclusion of RERP improves the distortion information capturing ability of all models in the DCN series, as opposed to their counterparts without RERP.
This observation is consistent with the results in Table~\ref{tab:ablation_components}.
The incorporation of both DAO and RERP in OOOPS showcases a remarkable distortion-aware modeling capability.

\section{Conclusion}
\label{conclusion}
In this work, we formulate the novel task of Open Panoramic Segmentation (OPS) for unfolding holistic scene understanding, which reaches Open FoV, Open Vocabulary, and Open Domain in surrounding scene segmentation.
We have designed the Deformable Adapter Network (DAN) and Random Equirectangular Projection (RERP) for enhancing distortion-aware segmentation capacities of the novel proposed OOOPS model.
Extensive experiments on WildPASS, Stanford2D3D, and Matterport3D benchmarks show that the proposed method significantly elevates the performance of state-of-the-art panoramic segmenters.

\section*{Acknowledgements}
This work was supported in part by the Ministry of Science, Research and the Arts of Baden-Württemberg (MWK) through the Cooperative Graduate School Accessibility through AI-based Assistive Technology (KATE) under Grant BW6-03, in part by BMBF through a fellowship within the IFI programme of DAAD, in part by the Helmholtz Association Initiative and Networking Fund on the HAICORE@KIT and HOREKA@KIT partition, in part by the National Key RD Program under Grant 2022YFB4701400, and in part by Hangzhou SurImage Technology Company Ltd.

%
%
\bibliographystyle{splncs04}
\bibliography{main}

\clearpage

\appendix

\section{Implementation Details}
\noindent\textbf{Hardware Setup.} In this work, we train our models using $4\times$A$40$ GPUs with respective $40$ GB memory. This computational node also contains $200$ GB CPU memory. The source code and the model checkpoints will be made publicly available.

\noindent\textbf{Training Settings.} 
The training process utilizes the AdamW optimizer, employing an initial learning rate of $0.0001$, a weight decay of $0.0001$, a batch size of $32$, and a total of $60,000$ training iterations.
During the training phase, the learning rate adheres to a polynomial schedule with a power of $0.9$.
Apart from the training specification listed in Table~\ref{tab:training_settings}, Table~\ref{tab:data_augmentation} illustrates the data augmentations and the corresponding parameters used in the training.
The short edge of an image is randomly resized in a range of $[320, 1024]$.
Afterward, the image is randomly cropped into $640{\times}640$.
For RandomBrightness, RandomContrast, RandomSaturation, and RandomHue, they are randomly applied with a probability of $0.5$.
The brightness delta, contrast range, saturation range, and hue delta of the aforementioned data augmentations are reported in Table~\ref{tab:data_augmentation}.
The input image is also flipped in the horizontal direction with $0.5$ probability before being forwarded to the model.
The whole training process takes roughly $10$ hours.
\begin{table}[h]
\vskip -4mm
\centering
\caption{Implementation details.}
\label{tab:implementation}
\vskip -4mm
\begin{minipage}{0.48\linewidth}
    \centering
    \subcaption{Training settings.}
    \label{tab:training_settings}
    \vskip -2ex
    \setlength{\tabcolsep}{10pt}
    \renewcommand{\arraystretch}{0.99}
    \resizebox{\columnwidth}{!}{
    \begin{tabular}{lc}
    \toprule[1.5pt]
    \textbf{Configurations} & \textbf{Parameter}  \\
    \midrule \midrule
    Optimizer             & AdamW                 \\
    Learning Rate         & $0.0001$ \\ 
    Weight Decay          & $0.0001$ \\ 
    Scheduler             & Poly. (power $0.9$)                \\
    Training Iterations   & $60,000$                    \\
    Batch Size per GPU    & $8$                    \\ 
    \bottomrule[1.5pt]
    \end{tabular}  
    }
\end{minipage}
\begin{minipage}{0.51\linewidth}
    \centering
    \subcaption{Data augmentation during the training process.}
    \label{tab:data_augmentation}
    \vskip -2ex
    \setlength{\tabcolsep}{28pt}
    \renewcommand{\arraystretch}{1.04}
    \resizebox{\columnwidth}{!}{
    \begin{tabular}{lc}
    \toprule[1.5pt]
    \textbf{Configurations} & \textbf{Parameter}  \\
    \midrule \midrule
    RandomResize              &     $[320, 1024]$           \\
    RandomCrop                &     $640 \times 640$          \\
    RandomBrightness          &     $32$          \\
    RandomContrast            &     $[0.5, 1.5]$          \\
    RandomSaturation          &     $[0.5, 1.5]$            \\
    RandomHue                 &     $18$            \\
    RandomFlip                &     Horizontal           \\
    \bottomrule[1.5pt]
    \end{tabular}
    }
\end{minipage}
\end{table}

\section{Qualitative Results}
\subsection{Visualization of Segmentation Predictions}
\begin{figure*}[h!]
    \vskip -2ex
    \centering
    \includegraphics[width=0.99\linewidth,height=5cm,trim=2 2 2 2,clip]{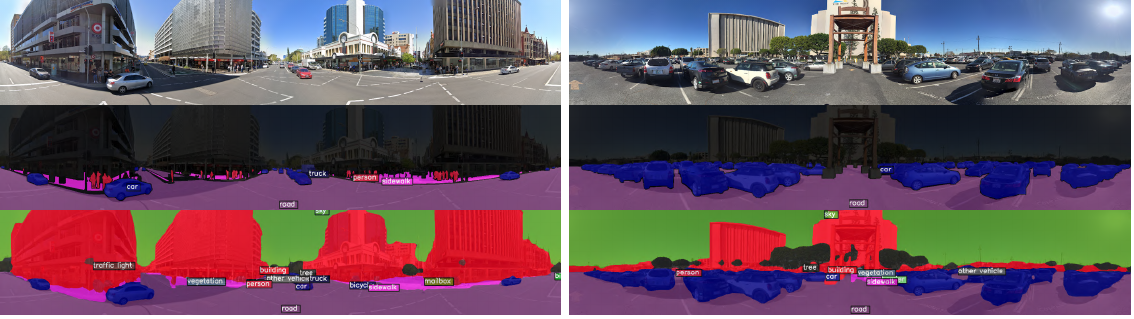}
    \caption{\textbf{Visualization on the WildPASS dataset}. First row: RGB images. Second row: close-vocabulary predictions of the proposed OOOPS model. Third row: open-vocabulary predictions of the proposed OOOPS model.}
    \label{fig_supp:wildpass}
\end{figure*}

\begin{figure*}[h!]
    \centering
    \includegraphics[width=0.99\linewidth,height=5cm,trim=2 2 2 2,clip]{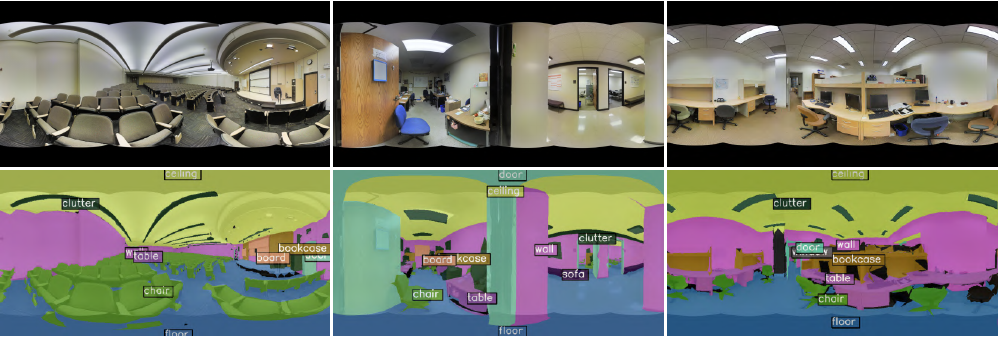}
    \caption{\textbf{Visualization on the Stanford2D3D dataset}. First row: RGB images. Second row: predictions of the proposed OOOPS model.}
    \label{fig_supp:s2d3d}
\end{figure*}

\begin{figure*}[h!]
    \centering
    \includegraphics[width=0.99\linewidth,height=5cm,trim=2 2 2 2,clip]{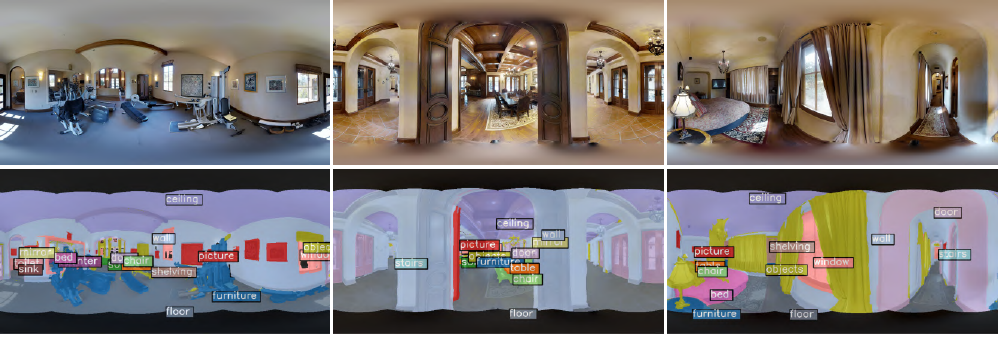}
    \caption{\textbf{Visualization on the Matterport3D dataset}. First row: RGB images. Second row: predictions of the proposed OOOPS model.}
    \label{fig_supp:mp3d}
    \vskip -4ex
\end{figure*}

Fig.~\ref{fig_supp:wildpass} presents the visualization of segmentation predictions on WildPASS~\cite{yang2021wildpass} dataset.
The first row is the input panoramic images.
The second row is the close-vocabulary segmentation predictions with only $8$ predefined categories in the WildPASS dataset while the last row is the open-vocabulary segmentation predictions with an arbitrary number of classes.
It can be observed that all pixels of the entire image have their own semantic meanings, showcasing the superiority of our proposed OOOPS model and the zero-shot learning ability.
It is worth noting that even the challenging category, \eg, mailbox, can be detected in the open-vocabulary setting.
Fig.~\ref{fig_supp:s2d3d} and Fig.~\ref{fig_supp:mp3d} illustrate the visualization of segmentation predictions on Stanford2D3D and Matterport3D datasets, respectively.
It's obvious that all predefined categories of these two datasets can be predicted correctly by the OOOPS model.
Beyond the correctness, the object deformations, \eg, the door in the middle column of Fig.~\ref{fig_supp:s2d3d}, can also be detected by our proposed OOOPS model, indicating the OOOPS model is aware of the image distortion and object deformation.

\subsection{Visualization of Deformable Offsets}
\begin{figure*}[h!]
    \vskip -2ex
    \centering
    \includegraphics[width=0.99\linewidth,height=5cm,trim=2 2 2 2,clip]{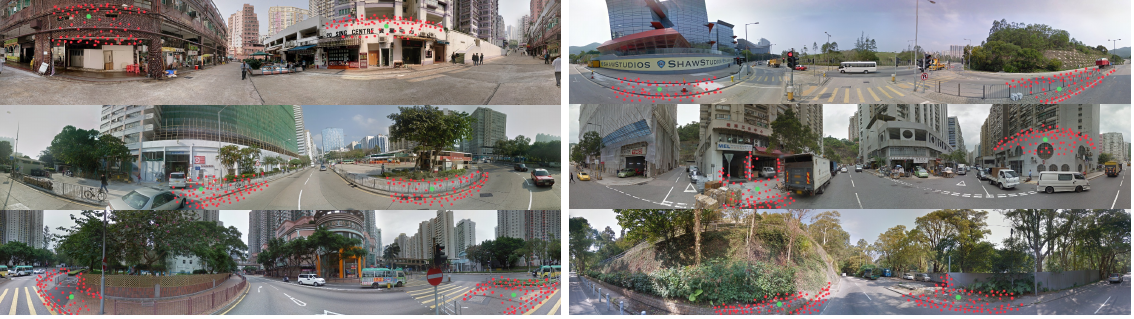}
    \caption{\textbf{Visualization of deformable offsets on WildPASS dataset.} The \textcolor{Green}{green points} \textcolor{Green}{$\bullet$} are the sample locations. The \textcolor{red}{red points} \textcolor{red}{$\bullet$} are deformable offsets in $2$ levels, indicating a deformable receptive field (\eg, each level has a $3\times3$ kernel size, resulting in $(3\times3)^2=81$ red points). Zoom in for a better view.}
    \label{fig_supp:wildpass-deformable}
\end{figure*}

\begin{figure*}[h!]
    \centering
    \includegraphics[width=0.99\linewidth,height=5cm,trim=2 2 2 2,clip]{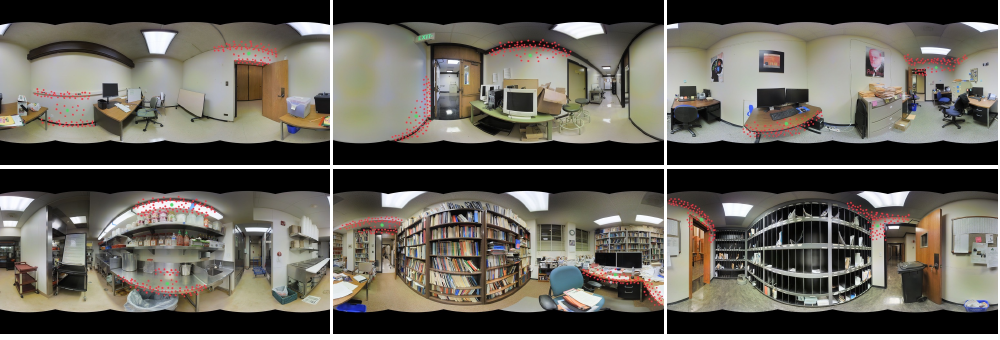}
    \caption{Visualization of deformable offsets on Stanford2D3D dataset.}
    \label{fig_supp:s2d3d-deformable}
\end{figure*}

\begin{figure*}[h!]
    \centering
    \includegraphics[width=0.99\linewidth,height=5cm,trim=2 2 2 2,clip]{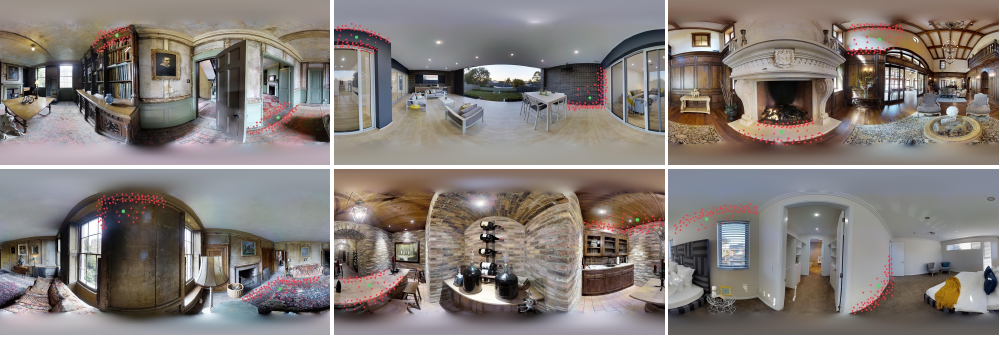}
    \caption{Visualization of deformable offsets on Matterport3D dataset.}
    \label{fig_supp:mp3d-deformable}
    \vskip -4ex
\end{figure*}

Since our OOOPS model is specifically designed for image distortion and object deformation, it is necessary to present the deformation-aware capability of the model.
Fig.~\ref{fig_supp:wildpass-deformable}, Fig.~\ref{fig_supp:s2d3d-deformable} and Fig.~\ref{fig_supp:mp3d-deformable} illustrate the deformable offsets on WilPASS, Stanford2D3D and Matterport3D dataset, respectively.
The \textcolor{Green}{green points} \textcolor{Green}{$\bullet$} are the sample locations.
The \textcolor{red}{red points} \textcolor{red}{$\bullet$} are deformable offsets in $2$ levels, indicating a deformable receptive field (\eg, each level has a $3\times3$ kernel size, resulting in $(3\times3)^2=81$ red points).
Leveraging the standard deviation of the cosine similarity vector calculated by the center pixel and all pixels within a kernel, the OOOPS model is capable of capturing the salient pixels, \eg, edge pixels of an image where strong panoramic distortion usually occurs.
For example, the sidewalk in the first row and second column of Fig.~\ref{fig_supp:wildpass-deformable} has a very strong distortion due to the Equirectangular Projection~\cite{ray2018erp} (ERP) from a globe to a plane resulting in a panoramic image.
Although the image distortion occurs, the green sample location has a deformable receptive field presented by red points along the edges of the sidewalk, indicating the deformation-aware capability of the OOOPS model.
The deformable awareness can be observed not only in the outdoor WildPASS panoramic dataset in Fig.~\ref{fig_supp:wildpass-deformable} but also in the indoor Stanford2D3D and Matterport3D datasets in Fig.~\ref{fig_supp:s2d3d-deformable} and Fig.~\ref{fig_supp:mp3d-deformable}.

\vskip -5ex

\vskip -5ex

\clearpage
\section{Ablation Study}
\subsection{Adaptive Shuffling Patches}
We conduct an experiment that divides pinhole images into parts of adaptive sizes when doing RERP augmentation, similar to Mosaic augmentation.
The results in Table~\ref{tab:supp_different_size_RERP} indicate adaptive sizes can further boost the model performance.
\begin{table}[t]
    \centering
    \caption{mIoU of RERP with adaptive sizes on WildPASS, Stanford2D3D, and Matterport3D datasets. mIoU is in percentage (\%).}
    \label{tab:supp_different_size_RERP}
    \renewcommand{\arraystretch}{0.5}
    \resizebox{0.85\columnwidth}{!}{
    \begin{tabular}{l c c c l}
        \toprule[1.5pt]
        \textbf{Method} & \textbf{WildPASS} & \textbf{Stanford2D3D} & \textbf{Matterport3D} \\ \midrule\midrule
        RERP & 58.0 & 41.1 & 31.2 \\
        \midrule
        RERP w/ adaptive sizes & \textbf{58.5} & \textbf{41.5} & \textbf{31.6} \\
        \bottomrule[1.5pt]
    \end{tabular}
    }
\end{table}

\subsection{Simple Horizontal Rotation}
The simple horizontal rotation~\cite{kim2023panoramic} is used to get a new panorama with a different viewpoint.
It is applied to panoramas, not pinhole images.
RERP is used to transform a pinhole image into a panorama-like image.
It is applied to pinhole images, not panoramas.
We experiment with applying simple horizontal rotation to pinhole images.
From Table~\ref{tab:supp_rotation} we find that simple horizontal rotation does not bring gains, 
which falls behind the one with RERP.
\begin{table}[t]
    \centering
    \caption{mIoU of RERP and simple horizontal rotation on WildPASS, Stanford2D3D, and Matterport3D datasets. mIoU is in percentage (\%).}
    \label{tab:supp_rotation}
    \renewcommand{\arraystretch}{0.5}
    \resizebox{0.85\columnwidth}{!}{
    \begin{tabular}{l c c c l}
        \toprule[1.5pt]
        \textbf{Method} & \textbf{WildPASS} & \textbf{Stanford2D3D} & \textbf{Matterport3D} \\ \midrule\midrule
        OOOPS w/o RERP & 57.0 & 39.5 & 31.1 \\
        \midrule
        OOOPS w/ Rotation & 57.0 & 39.5 & 31.1 \\
        \midrule
        OOOPS w/ RERP & \textbf{58.0} & \textbf{41.1} & \textbf{31.2} \\
        \bottomrule[1.5pt]
    \end{tabular}
    }
\end{table}

\section{Discussion}
\noindent\textbf{Limitations and Future Work.}
In this work, we focus on the open panoramic segmentation, where the models are trained in the narrow-FoV pinhole source domain in an open-vocabulary setting while evaluated in the wide-FoV panoramic target domain.
Compared to the state-of-the-art methods trained in a close-vocabulary setting, the limitations of the proposed OOOPS model are obvious.
The performance of the open-vocabulary model falls short of the close-vocabulary ones.
The architectural design does not encompass the 360{\textdegree} boundaries of panoramas, providing an opportunity for improving seamless scene segmentation.
Additionally, the generalization capability of the OOOPS model can be evaluated using surround-view fisheye images.
Our future plans involve extending the proposed solution to encompass panoramic panoptic segmentation.

\noindent\textbf{Societal Impacts.}
The proposed Open Panoramic Segmentation (OPS) task and the OOOPS model with Random Equirectangular Projection (RERP) enable distortion-aware open-vocabulary panoramic semantic segmentation in different open domains even though there are no training-sufficient dense-annotated panoramic labels.
Evidently, this represents a great technological advancement that necessitates not only strategic utilization but also a thorough awareness of the inherent risks.
Although this work is able to predict an arbitrary number of classes in holistic scene understanding, \eg, in the autonomous driving scenario regarding panoramic distortion, the performance of the model should be further improved for the safety of both drivers and pedestrians.
Apart from the outdoor applications, the indoor navigation of robots using panoramas is supposed to focus more on open-vocabulary prediction correctness so that robots can better serve humans, avoiding unexpected potential dangers.

\end{document}